\documentclass[11pt,a4paper]{article}
\usepackage{times,latexsym}
\usepackage{url}
\usepackage[T1]{fontenc}

\usepackage[acceptedWithA]{tacl2021v1}

\usepackage{amsmath}
\usepackage{amssymb}
\usepackage{graphicx}
\usepackage{bm}
\usepackage{multirow}
\usepackage{booktabs}
\usepackage[inline,shortlabels]{enumitem}
\newcommand{\vctr}[1]{\bm{#1}}
\newcommand{\mtrx}[1]{\bm{#1}}
\newcommand{\bettertextcircled}[1]{\raisebox{.5pt}{\textcircled{\raisebox{-.9pt} {#1}}}}
\newcommand\smallmath[2]{#1{\raisebox{\dimexpr \fontdimen 22 \textfont 2
      - \fontdimen 22 \scriptscriptfont 2 \relax}{$\scriptscriptstyle #2$}}}
\newcommand\smalloplus{\smallmath\mathbin\oplus}
\newcommand\smallominus{\smallmath\mathbin\ominus}

\usepackage{xspace,mfirstuc,tabulary}

\newif\iftaclinstructions
\taclinstructionsfalse 
\iftaclinstructions
\renewcommand{\confidential}{}
\renewcommand{\anonsubtext}{(No author info supplied here, for consistency with
TACL-submission anonymization requirements)}
\newcommand{\instr}
\fi

\iftaclpubformat 

\else

\fi

\title{Robust Dialogue State Tracking with Weak Supervision and Sparse Data}

\author{Michael Heck, Nurul Lubis, Carel van Niekerk,\\\textbf{Shutong Feng, Christian Geishauser, Hsien-Chin Lin, Milica Ga\v{s}i\'{c}} \\
  Heinrich Heine University Düsseldorf, Germany \\
  \texttt{\{heckmi,lubis,niekerk,fengs,geishaus,linh,gasic\}@hhu.de}}

\date{}

\begin{document}
\maketitle
\begin{abstract}
Generalising dialogue state tracking (DST) to new data is especially challenging due to the strong reliance on abundant and fine-grained supervision during training. Sample sparsity, distributional shift and the occurrence of new concepts and topics frequently lead to severe performance degradation during inference.
In this paper we propose a training strategy to build extractive DST models without the need for fine-grained manual span labels. Two novel input-level dropout methods mitigate the negative impact of sample sparsity. We propose a new model architecture with a unified encoder that supports value as well as slot independence by leveraging the attention mechanism. We combine the strengths of triple copy strategy DST and value matching to benefit from complementary predictions without violating the principle of ontology independence.
Our experiments demonstrate that an extractive DST model can be trained without manual span labels. Our architecture and training strategies improve robustness towards sample sparsity, new concepts and topics, leading to state-of-the-art performance on a range of benchmarks. We further highlight our model's ability to effectively learn from non-dialogue data.
\end{abstract}

\section{Introduction}

Generalisation and robustness are among the key requirements for naturalistic conversational abilities of task-oriented dialogue systems~\cite{edlund2008towards}.
In a dialogue system, dialogue state tracking (DST) solves the task of extracting meaning and intent from the user input, and keeps track of the user's goal over the continuation of a conversation
as part of a dialogue state (DS)~\cite{young2010hidden}.
A recommendation and booking system for places, for instance, needs to gather user preferences in terms of budget, location, etc. Concepts like these are assembled in an ontology on levels of domain (e.g., \emph{restaurant} or \emph{hotel}), slot (e.g., \emph{price} or \emph{location}), and value (e.g., ``expensive'' or ``south'').
Accurate DST is vital to a robust dialogue system, as the system's future actions depend on the conversation's current estimated state.
However, generalising DST to new data and domains is especially challenging. The reason is the strong reliance on supervised training.

Virtually all top-performing DST methods either entirely or partially extract values directly from context~\cite{ni2021recent}.
However, training these models robustly is a demanding task. Extractive methods usually rely on fine-grained labels on word level indicating the precise locations of value mentions. Given the richness of human language and the ability to express the same canonical value in many different ways,
producing such labels is challenging and very costly, and it is no surprise that datasets of such kind are rare~\cite{zhang2020recent,deriu2021survey}.
Reliance on detailed labels has another downside; datasets are usually severely limited in size. This in turn leads to the problem of sample sparsity, which increases the risk for models to over-fit to the training data, for instance by memorising values in their respective contexts. Over-fitting prevents a state tracker to generalise to new contexts and values, which is likely to break a dialogue system entirely~\cite{qian2021annotation}.
Recently, domain-independent architectures have been encouraged to develop systems that may be built once, and then applied to new scenarios with no or little additional training~\cite{rastogi2020towards,rastogi2020schema}. However, training such flexible models robustly remains a challenge, and the ever-growing need for more training samples spurs creativity to leverage non-dialogue data~\cite{heck2020out, namazifar2021language}. 

We propose novel strategies for extractive DST that address the following four issues of robustness and generalisation.
(1) We solve \emph{the problem of requiring fine-grained span labels} with a self-supervised training scheme.
Specifically, we learn from random self-labeled samples how to locate occurrences of arbitrary values.
All that is needed for training a full DST model is the dialogue state ground truth, which is undoubtedly much easier to obtain than fine-grained span labels.
(2) We handle \emph{the sample sparsity problem} by introducing two new forms of input-level dropout into training. Our proposed dropout methods are easy to apply and provide a more economical alternative to data augmentation to prevent memorisation and over-fitting to certain conversation styles or dialogue patterns.
(3) We add a value matching mechanism on top of extraction to enhance \emph{robustness towards previously unseen concepts}. Our value matching is entirely optional and may be utilised if a set of candidate values is known during inference, for instance from a schema or API.
(4) We propose a new architecture that is entirely domain-agnostic to facilitate \emph{transfer to unseen slots and domains}. For that, our model relies on the attention mechanism and conditioning on natural language slot descriptions. The established slot-independence enables zero-shot transfer. We will demonstrate that we can actively teach to track new domains by learning from non-dialogue data. This is non-trivial as the model must learn to interpret dialogue data from exposure to unstructured data.

\begin{figure*}[t]
	\centering
	\includegraphics[page=2, trim=0.0cm 3.3cm 0.0cm 2.9cm, clip=true, width=1.00\linewidth,]{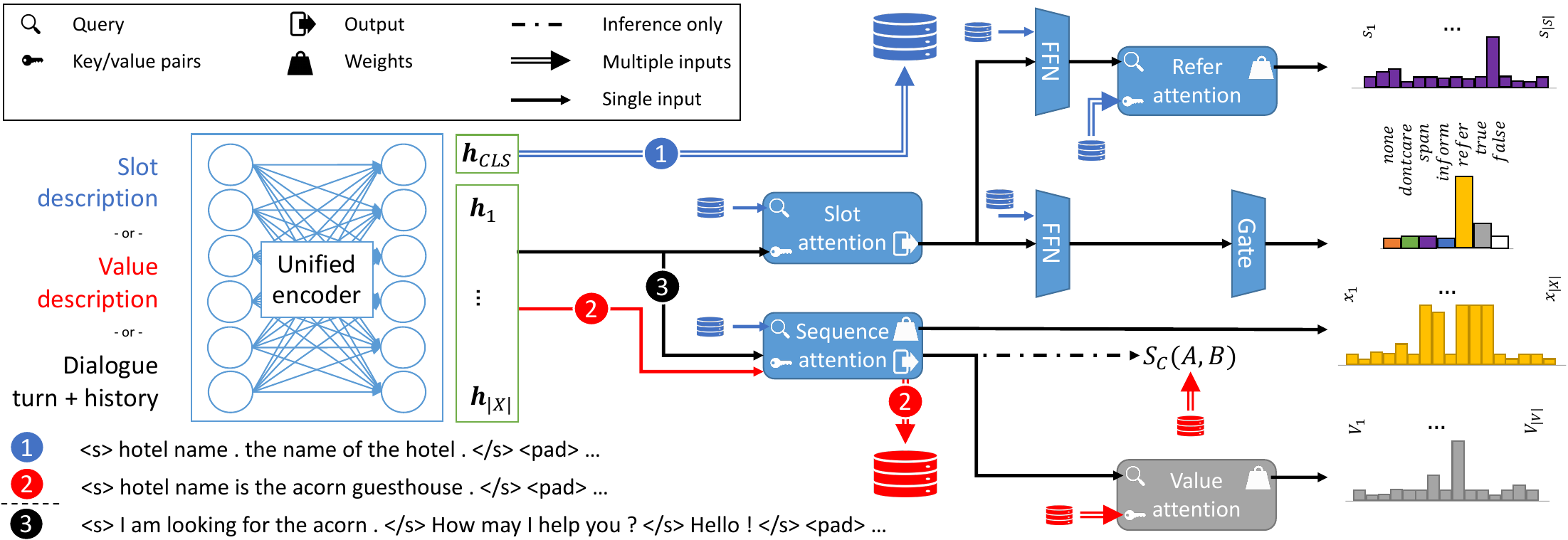}
	\caption{Proposed model architecture. TripPy-R takes the turn and dialogue history as input and outputs a DS. All inputs are encoded separately with the same fine-tuned encoder. For inference, slot and value representations are encoded once and then stored in databases for retrieval.}
	\label{fig:model}
	\vspace{-12pt}
\end{figure*}

\section{Related Work}
\label{sec:related}

Traditional DS trackers perform prediction over a fixed ontology~\cite{mrkvsic2016neural,liu2017end,zhong2018global}
and therefore have various limitations in more complex scenarios~\cite{ren2018towards,nouri2018toward}. The idea of fixed ontologies is not sustainable for real world applications, as new concepts become impossible to capture during test time.
Moreover, the demand for finely labeled data quickly grows with the ontology size, causing scalability issues.

Recent approaches to DST
extract values directly from the dialogue context via span prediction~\cite{xu2018end,gao2019dialog,chao2019bert}, removing the need for fixed value candidate lists. An alternative to this mechanism is value generation via soft-gated pointer-generator copying~\cite{wu2019transferable, kumar2020ma, kim2019efficient}. Extractive methods have limitations as well, since many values may be expressed variably or implicitly. Contextual models such as BERT~\cite{devlin2018bert} support generalisation over value variations to some extent~\cite{lee2019sumbt, chao2019bert, gao2019dialog}, and hybrid approaches try to mitigate the issue by resorting to picklists~\cite{zhang2019find}.

TripPy~\cite{heck-etal-2020-trippy} jointly addresses the issues of coreference, implicit choice and value independence with a triple copy strategy.
Here, a Transformer-based~\cite{vaswani2017attention} encoder projects each dialogue turn into a semantic embedding space. Domain-slot specific slot gates then decide whether or not a slot-value is present in the current turn in order to update the dialogue state. In case of presence, the slot gates also decide which of the following three copy mechanisms to use for extraction. (1) Span prediction extracts a value directly from input. For that, domain-slot specific span prediction heads predict per token whether it is the beginning or end of a slot-value. (2) Informed value prediction copies a value from the list of values that the system informed about. This solves the implicit choice issue, where the user might positively but implicitly refer to information that the system provided. (3) Coreference prediction identifies cases where the user refers to a value that has already been assigned to a slot earlier and should now also be assigned to another slot in question.
TripPy shows good robustness towards new data from known domains since it does not rely on a priori knowledge of value candidates. However, it does not support transfer to new topics, since the architecture is ontology specific. Transfer to new domains or slots is therefore impossible without re-building the model. TripPy also ignores potentially available knowledge about value candidates, since its copy mechanisms operate solely on the input. Lastly, training requires fine-grained span labels, complicating the transfer to new datasets.

While contemporary approaches to DST leverage parameter sharing and transfer learning~\cite{rastogi2020schema, lin2021zero}, the need for finely-labeled training data is still high. Sample sparsity often causes model biases in the form of memorisation or other types of over-fitting. Strategies to appease the hunger of larger models are the exploitation of out-of-domain dialogue data for transfer effects~\cite{wu2020tod} and data augmentation~\cite{campagna2020zero, yu2020score, li2020coco, dai2021preview}.
However, out-of-domain dialogue data is limited in quantity as well. Data augmentation still requires high level knowledge about dialogue structures and an adequate data generation strategy. Ultimately, more data also means longer training.
We are aware of only one recent work that attempts DST with weak supervision. \citet{liang2021attention} take a few-shot learning approach using only a subset of fully labeled training samples---typically from the end of conversations---to train a soft-gated pointer-generator network. In contrast, with our approach to spanless training, we reduce the level of granularity needed for labels to train extractive models. Note that these strategies are orthogonal.

\section{TripPy-R: Robust Triple Copy DST}

Let $\{(U_1, M_1), \dots, (U_T, M_T)\}$ be the sequence of turns that form a dialogue. $U_t$ and $M_t$ are the token sequences of the user utterance and preceding system utterance at turn $t$. The task of DST is (1) to determine for every turn whether any of the domain-slot pairs in $S = \{S_1, \dots, S_N\}$ is present, (2) to predict the values for each $S_n$ and (3) to track the dialogue state $\mathit{DS}_t$.
Our starting point is triple copy strategy DST~\cite{heck-etal-2020-trippy}, because it has already been designed for robustness towards unseen values.
However, we propose a new architecture with considerable differences to the baseline regarding its design, training and inference to overcome the drawbacks of previous approaches as laid out in Section~\ref{sec:related}.
We call our proposed framework \emph{TripPy-R} (pronounced ``trippier''), \textbf{R}obust \textbf{tri}ple co\textbf{py} strategy DST\footnote{\url{https://gitlab.cs.uni-duesseldorf.de/general/dsml/trippy-r-public}}. Figure~\ref{fig:model} is a depiction of our proposed model.

\subsection{Model Layout}

\paragraph{Joint Components}

We design our model to be entirely domain-agnostic, adopting the idea of conditioning the model with natural language descriptions of concepts~\cite{bapna2017towards, rastogi2020towards}. For that, we use data-independent prediction heads that can be conditioned with slot descriptions to solve the tasks required for DST. This is different to related work such as~\citet{heck-etal-2020-trippy}, which uses data-dependent prediction heads whose number depends on the ontology size.
In contrast, prediction heads in TripPy-R are realised via the attention mechanism~\cite{DBLP:journals/corr/BahdanauCB14}.
Specifically, we use scaled dot-product attention, implemented as multi-head attention according to and defined by~\citet{vaswani2017attention}.
We utilise this mechanism to query the input for the presence of information.
Among other things,
we deploy attention to predict whether or not a slot-value is present in the input, or to conduct sequence tagging---rather than span prediction---by assigning importance weights to input tokens.

\paragraph{Unified Context/Concept Encoder}

Different from other domain-agnostic architectures~\cite{lee2019sumbt,ma2019end}, we rely on a single encoder that is shared among encoding tasks. This unified encoder is used to produce representations for dialogue turns and natural language slot and value descriptions.
The encoder function is
$\mathrm{Enc}(X) = [\vctr{h}_{\mathrm{CLS}}, \vctr{h}_1, \dots, \vctr{h}_{\mathrm{|X|}}]$,
where $X$ is a sequence of input tokens. $\vctr{h}_{\mathrm{CLS}}$ can be interpreted as a representation of the entire input sequence. The vectors $\vctr{h}_1$ to $\vctr{h}_{\mathrm{|X|}}$ are contextual representations for the sequence of input tokens. We define
$\mathrm{Enc_P}(X) = [\vctr{h}_{\mathrm{CLS}}]$ and
$\mathrm{Enc_S}(X) = [\vctr{h}_1, \dots, \vctr{h}_{\mathrm{|X|}}]$
as the pooled encoding and sequence encoding of $X$, respectively.

Dialogue turns and natural language slot and value descriptions are encoded as
\begin{equation*}
\begin{split}
    \mtrx{R}_t =& \mathrm{Enc_S}(x_{\mathrm{CLS}} \oplus U_t \oplus x_{\mathrm{SEP}} \oplus M_t \oplus \\
                        &\hspace{28pt}x_{\mathrm{SEP}} \oplus H_{t} \oplus x_{\mathrm{SEP}}), \\
    \vctr{r}_{S_i} =& \mathrm{Enc_P}(x_{\mathrm{CLS}} \oplus S_i \oplus \mathrm{"."} \oplus S_i^\mathrm{desc} \oplus x_{\mathrm{SEP}}),\\
    \mtrx{R}_{V_{S_i,j}} =& \mathrm{Enc_S}(x_{\mathrm{CLS}} \oplus S_i \oplus \mathrm{"is"} \oplus V_{S_i,j} \oplus x_{\mathrm{SEP}}),
\end{split}
\label{eq:enc}
\end{equation*}
where $H_t = \{(U_{t-1}, M_{t-1}), \dots, (U_1, M_1)\}$ is the history of the dialogue up to turn $t$. The special token $x_{\mathrm{CLS}}$ initiates every input sequence, and $x_{\mathrm{SEP}}$ is a separator token to provide structure to multi-sequence inputs.
$S_i^\mathrm{desc}$ is the slot description of slot $S_i$ and $V_{S_i,j}$ is a candidate value $j$ for slot $S_i$.

\paragraph{Conditioned Slot Gate}

The slot gate outputs a probability distribution over the output classes $C =$ $\{\mathit{none}$, $\mathit{dontcare}$, $\mathit{span}$, $\mathit{inform}$, $\mathit{refer}$, $\mathit{true}$, $\mathit{false}\}$.
Our slot gate can be conditioned to perform a prediction for one particular slot, allowing our architecture to be ontology independent.
The \emph{slot attention} layer attends to token representations of a dialogue turn given the representation of a particular slot $S_i$ as query, i.e.,
\begin{equation}
\begin{split}
    [\vctr{g}_{\mathrm{o}}, \vctr{g}_{\mathrm{w}}] =& \mathrm{MHA}_{\mathrm{g}}(\vctr{r}_{S_i}, \mtrx{R}_{t}, \mtrx{R}_{t}),
\end{split}
\label{eq:slot_att}
\end{equation}
where $\mathrm{MHA}_{(\cdot)}(\mtrx{Q}, \mtrx{K}, \mtrx{V}, \hat{\vctr{k}})$ is a multi-head attention layer
that expects a query matrix $\mtrx{Q}$, a key matrix $\mtrx{K}$, a value matrix $\mtrx{V}$ and an optional masking parameter $\hat{\vctr{k}}$. $\vctr{g}_{\mathrm{o}}$ is the layer-normalised~\cite{ba2016layer} attention output and $\vctr{g}_{\mathrm{w}}$ are the attention weights.
For classification, the attention output is piped into a feed-forward network (FFN) conditioned with $S_i$,
\begin{equation*}
    \vctr{g}_{\mathrm{s}} = \mathrm{softmax}(\mathrm{L}_3(\mathrm{G}_2(\vctr{r}_{S_i} \oplus \mathrm{G}_1(\vctr{g}_{\mathrm{o}})))) \in \mathbb{R}^7,
\end{equation*}
where $\mathrm{L}_{(\cdot)}(\vctr{x}) = \mtrx{W}_{(\cdot)} \cdot \vctr{x} + \vctr{b}_{(\cdot)}$ is a linear layer, and $\mathrm{G}_{(\cdot)}(\vctr{x}) = \mathrm{GeLU}(\mathrm{L}_{(\cdot)}(\vctr{x}))$~\cite{hendrycks2016gaussian}.

\paragraph{Sequence Tagging}

In order to keep the value extraction directly from the input ontology-independent as well, our model re-purposes attention to perform sequence tagging.
If the slot gate predicts $\mathit{span}$, the \emph{sequence attention} layer attends to token representations of the current dialogue turn given $\vctr{r}_{S_i}$ as query, analogous to Eq.~(\ref{eq:slot_att}):
\begin{equation}
    [\vctr{q}_{\mathrm{o}}, \vctr{q}_{\mathrm{w}}] = \mathrm{MHA}_{\mathrm{q}}(\vctr{r}_{S_i}, \mtrx{R}_{t}, \mtrx{R}_{t}, \hat{\vctr{r}}_{t}).
\label{eq:seq_att}
\end{equation}
Here,
$\hat{\vctr{r}}_{t}$ is an input mask that only allows attending to representations of user utterances.

In contrast to other work that leverages attention for DST~\cite{lee2019sumbt,wu2019transferable}, we explicitly teach the model \emph{where} to put the attention. This way, the predicted attention weights $\vctr{q}_{\mathrm{w}}$ become the sequence tagging predictions. Tokens that belong to a value are assigned a weight of 1, all other tokens are weighted 0. Since $\|\vctr{q}_{\mathrm{w}}\|_1 = 1$, we scale the target label sequences during training. During inference, we normalise $\vctr{q}_{\mathrm{w}}$, i.e.,
\begin{equation}
  \hat{\vctr{q}}_{\mathrm{w}} = [\hat{q}_1,\dots,\hat{q}_{|X|}], \text{with } 
  \hat{q}_j = \frac{q_{\mathrm{w},j} - \frac{1}{|X|}}{\max_{\forall q \in \vctr{q}_{\mathrm{w}}} q},
  \label{eq:seq_att_quant}
\end{equation}
so that we can infer sequence tags according to an ``\emph{IO}'' tagging scheme~\cite{ramshaw-marcus-1995-text}.
All $\hat{q}_j > 0$ are assigned the ``\emph{I}'' tag, all others the ``\emph{O}'' tag.
The advantage of sequence tagging over span prediction is that training can be performed using labels for multiple occurrences of the same slot-value in the input---for instance in the current turn and the dialogue history---, and that multiple regions of interest can be predicted.
To extract a value from the context, we pick the sequence with the highest average token weight according to $\hat{\vctr{q}}_{\mathrm{w}}$ among all sequences of tokens that were assigned the ``\emph{I}'' tag and denote this value prediction as $\mathrm{Val}(\hat{\vctr{q}}_{\mathrm{w}})$.

\paragraph{Informed Value Prediction}

We adopt informed value prediction from TripPy. Ontology independence is established via our conditioned slot gate.
The inform memory $I_t = \{I_t^1, \dots, I_t^{|S|}\}$ tracks slot-values that were informed by the system in the current dialogue turn $t$. If the user positively refers to an informed value, and if the user does not express the value such that sequence tagging can be used---i.e., the slot gate predicts $\mathit{inform}$---, then the value ought to be copied from $I_t$ to $\mathit{DS}_t$.

We know from works on cognition that ``all collective actions are built on common ground and its accumulation''~\cite{clark1991grounding}. In other words, it must be established in a conversation what has been understood by all participants. The process of forming mutual understanding is known as \emph{grounding}. Informed value prediction in TripPy-R serves as a grounding component. As long as the information shared by the system has not yet been grounded, i.e., confirmed by the user, it is not added to the DS. This is in line with information state and dialogue management theories such as devised by~\citet{larsson2000information}, which view grounding as essential to the theory of information states and therefore DST.

\paragraph{Coreference Prediction}

Although TripPy supports coreference resolution, this mechanism is limited to an a priori known set of slots. We use attention to establish slot independence for coreference resolution to overcome this limitation. If the slot gate predicts $\mathit{refer}$ for a slot $S_i$, i.e., that it refers to a value that has previously been assigned to another slot, then the \emph{refer attention} needs to predict the identity of said slot $S_j$, i.e.,
\begin{equation*}
    [\vctr{f}_{\mathrm{o}}, \vctr{f}_{\mathrm{w}}] = \mathrm{MHA}_{\mathrm{f}}(\mathrm{G}_5(\vctr{r}_{S_i} \oplus \mathrm{G}_4(\vctr{g}_{\mathrm{o}})), \mtrx{R}_{S}, \mtrx{R}_{S}),
\end{equation*}
where the slot attention output $\vctr{g}_{\mathrm{o}}$ is first piped through an FFN. $\mtrx{R}_{S} = [\vctr{r}_{S_1},\dots,\vctr{r}_{S_{|S|}}] \in \mathbb{R}^{d \times |S|}$ is the matrix of stacked slot representations and $\vctr{f}_{\mathrm{w}}$ is the set of weights assigned to all candidate slots for $S_j$. The slot with the highest assigned weight is then our referred slot $S_j$. To resolve a coreference, $S_i$ is updated with the value of $S_j$. During inference, $\mtrx{R}_{S}$ can be modified as desired to accommodate new slots.

\paragraph{Value Matching}

In contrast to picklist based methods such as by~\citet{zhang2019find}, TripPy-R performs value matching as an optional step.
We first create slot-value representations for all value candidates $V_{S_i,j}$ of slot $S_i$, and learn matching of dialogue context $\vctr{q}_{\mathrm{o}}$ to the list of candidate values via \emph{value attention}:
\begin{equation}
\begin{split}
    [\vctr{r}_{V_{S_i,j}}, \vctr{v}_{\mathrm{w}}] &= \mathrm{MHA}_{\mathrm{q}}(\vctr{r}_{S_i}, \mtrx{R}_{V_{S_i,j}}, \mtrx{R}_{V_{S_i,j}}), \\
    [\vctr{m}_{\mathrm{o}}, \vctr{m}_{\mathrm{w}}] &= \mathrm{MHA}_{\mathrm{m}}(\vctr{q}_{\mathrm{o}}, \mtrx{R}_{V_{S_i}}, \mtrx{R}_{V_{S_i}}).
\end{split}
\label{eq:seq_att_for_val}
\end{equation}
where $\mtrx{R}_{V_{S_i}} = [\vctr{r}_{V_{S_i,1}},\dots,\vctr{r}_{V_{S_i,|V_{S_i}|}}] \in \mathbb{R}^{d \times |V_{S_i}|}$.
$\vctr{m}_{\mathrm{w}}$ should place a weight close to 1 on the correct value and weights close to 0 on all the others.
Dot-product attention as used in our model is defined as $\mathrm{softmax}(\mtrx{Q} \cdot \mtrx{K}^\top) \cdot \mtrx{V}$.
Computing the dot product between input and candidate value representations is proportional to computing their cosine similarities, which is
$\mathrm{cos}(\theta) = \frac{\vctr{q} \cdot \vctr{k}}{\|\vctr{q}\| \cdot \|\vctr{k}\|}$ $\forall \vctr{q} \in \mtrx{Q}, \vctr{k} \in \mtrx{K}$.
Therefore, optimising the model to put maximum weight on the correct value and to minimise the weights on all other candidates forces representations of the input and of values occurring in that input to be closer in their common space, and vice versa.

\subsection{Training and Inference}
\label{sec:trippyr:ssec:training}

Each training step requires the dialogue turn and all slot and value descriptions be encoded. Our unified encoder re-encodes all slot descriptions at each step. Since the number of values might be in the range of thousands, we encode them once for each epoch. The encoder is fine-tuned towards encoding all three input types.
We optimise our model given the joint loss for each turn,
\begin{equation}
    \mathcal{L} = \lambda_\mathrm{g} \cdot \mathcal{L}_{\mathrm{g}} + \lambda_\mathrm{q} \cdot \mathcal{L}_{\mathrm{q}} + \lambda_\mathrm{f} \cdot \mathcal{L}_{\mathrm{f}} + \lambda_\mathrm{m} \cdot \mathcal{L}_{\mathrm{m}}, \\
  \label{eq:loss}
\end{equation}
\begin{align*}
    \mathcal{L}_{\mathrm{g}} &= \Sigma_i\ell(\vctr{g}_{\mathrm{s}}, L^\mathrm{g}_{S_i}), &&L^\mathrm{g}_{S_i} \in C, \\
    \mathcal{L}_{\mathrm{q}} &= \Sigma_i\ell(\vctr{q}_{\mathrm{w}}, \vctr{l}^\mathrm{q}_{S_i}/\|\vctr{l}^\mathrm{q}_{S_i}\|_1), &&\vctr{l}^\mathrm{q}_{S_i} \in \{0, 1\}^{|X|}, \\
    \mathcal{L}_{\mathrm{f}} &= \Sigma_i\ell(\vctr{f}_{\mathrm{w}}, \vctr{l}^\mathrm{f}_{S_i}), &&\vctr{l}^\mathrm{f}_{S_i} \in \{0, 1\}^{|S|}, \\
    \mathcal{L}_{\mathrm{m}} &= \Sigma_i\ell(\vctr{m}_{\mathrm{w}}, \vctr{l}^\mathrm{m}_{S_i}), &&\vctr{l}^\mathrm{m}_{S_i} \in \{0, 1\}^{|V_{S_i}|}.
\end{align*}
Here, $\ell(\cdot, \cdot)$ is the loss between a prediction and a ground truth. $\mathcal{L}_{\mathrm{g}}$, $\mathcal{L}_{\mathrm{q}}$, $\mathcal{L}_{\mathrm{f}}$ and $\mathcal{L}_{\mathrm{m}}$ are the joint losses of the slot gate, sequence tagger, coreference prediction and value matching.
It is $\|\cdot\|_1 = 1$ for $\vctr{l}^\mathrm{f}_{S_i}$ and $\vctr{l}^\mathrm{m}_{S_i}$, i.e., labels for coreference prediction and value matching are 1-hot vectors.
Back-propagating $\mathcal{L}_{\mathrm{m}}$ also affects the sequence tagger.
We scale $\vctr{l}^\mathrm{q}_{S_i}$ since sequence tagging may have to label more than one token as being part of a value.

During inference, the model can draw from the rich output of the model, i.e., slot gate predictions, coreference prediction, sequence tagging and value matching to adequately update the dialogue state.
Slot and value descriptions are encoded only once with the fine-tuned encoder, then stored in databases as illustrated in Figure~\ref{fig:model} in steps \bettertextcircled{1} and \bettertextcircled{2}. Pre-encoded slots condition the attention and FFN layers, and pre-encoded values are used for value matching. Note that it is straightforward to update these databases on-the-fly for a running system, thus easily expanding its capacities. Step \bettertextcircled{3} is the processing of dialogue turns to perform dialogue state update prediction.

\subsection{Dialogue State Update}

At turn $t$, the slot gate predicts for each slot $S_i$ how it should be updated. $none$ means that no update is needed. $\mathit{dontcare}$ denotes that any value is acceptable to the user. $\mathit{span}$ indicates that a value is extractable from any of the the user utterances $\{U_{t},\dots,U_1\}$. $\mathit{inform}$ denotes that the user refers to a value uttered by the system in $M_t$. $\mathit{refer}$ indicates that the user refers to a value that is already present in $\mathit{DS}_t$ in a different slot. Classes $\mathit{true}$ and $\mathit{false}$ are used by slots that take binary values.

If candidate values are known at inference, TripPy-R can utilise value matching to benefit from supporting predictions for the $\mathit{span}$ case. Since sequence tagging and value matching predictions would compete over the slot update, we use confidence scores to make an informed decision.
Given the current input, and candidate values for a slot, we can use the attention weights $\vctr{m}_{\mathrm{w}}$ of the value attention as individual scores for each value. We can also use the L2-norm between input and values, i.e., $e_{S_i,j} = \|\vctr{q}_{\mathrm{o}} - \vctr{r}_{V_{S_i,j}}\|_2$, and $\vctr{e}_{S_i} = [e_{S_i,1},\dots,e_{S_i,|V_{S_i}|}]$ is the score set.
Then
\begin{equation*}
  \mathrm{Conf}(C) = 1 - \frac{\min_{\forall c \in C} c}{\left( \left( \sum_{c \in C} c \right) - \min_{\forall c \in C} c \right) / |C|},
\end{equation*}
is applied to $\vctr{m}_{\mathrm{w}}$ and $\vctr{e}_{S_i}$ (interpreting them as multisets rather than vectors) to compute two confidence scores $\mathrm{Conf}(\vctr{m}_{\mathrm{w}})$ and $\mathrm{Conf}(\vctr{e}_{S_i})$ for the most likely value candidate. This type of confidence captures the notion of difference between the best score and the mean of all other scores, intuitively expressing model certainty.
$\mathrm{Val}(\vctr{m}_{\mathrm{w}}) = \mathrm{argmax}(\vctr{m}_{\mathrm{w}})$ and $\mathrm{Val}(\vctr{e}_{S_i}) = \mathrm{argmax}(\vctr{e}_{S_i})$ are the most likely candidates according to value attention and L2-norm.
For any slot that was predicted as $\mathit{span}$, the final prediction is
\begin{equation*}
    S_i^* = 
    \begin{cases}
        \mathrm{Val}(\vctr{m}_{\mathrm{w}}), \text{if} &S_i \text{ is categorical\footnotemark } \land \\ 
        &\mathrm{Conf}(\vctr{m}_{\mathrm{w}}) > \tau \\
        \mathrm{Val}(\vctr{e}_{S_i}), \text{else if} &\mathrm{Conf}(\vctr{e}_{S_i}) > \tau \\
        \mathrm{Val}(\hat{\vctr{q}}_{\mathrm{w}}), \text{else},&
    \end{cases}
\end{equation*}
where $\tau$ $\in$ $[0,1]$ is a threshold parameter that controls the level of the model's confidence needed to still consider its value matching predictions.
\footnotetext{For the distinction of categorical and non-categorical slots, see~\citet{rastogi2020towards} and~\citet{zang2020multiwoz}.}

\section{Levels of Robustness in DST}

We propose the following methods to improve robustness in DST on multiple levels.

\subsection{Robustness to Spanless Labels}
\label{sec:robustness:ssec:spanless}

Our framework introduces a novel training scheme to learn from data without span labels, therefore lowering the demand for fine-grained labels.
We teach a proto-DST model that uses parts of TripPy-R's architecture to tag random token sub-sequences that occur in the textual input.
We use this model to locate value occurrences in each turn $t$ of a dialogue as listed in the labels for $\mathit{DS}_t$.

\begin{figure}[t]
	\centering
	\includegraphics[page=1, trim=0.0cm 10cm 15cm 0.0cm, clip=true, width=1.00\linewidth,]{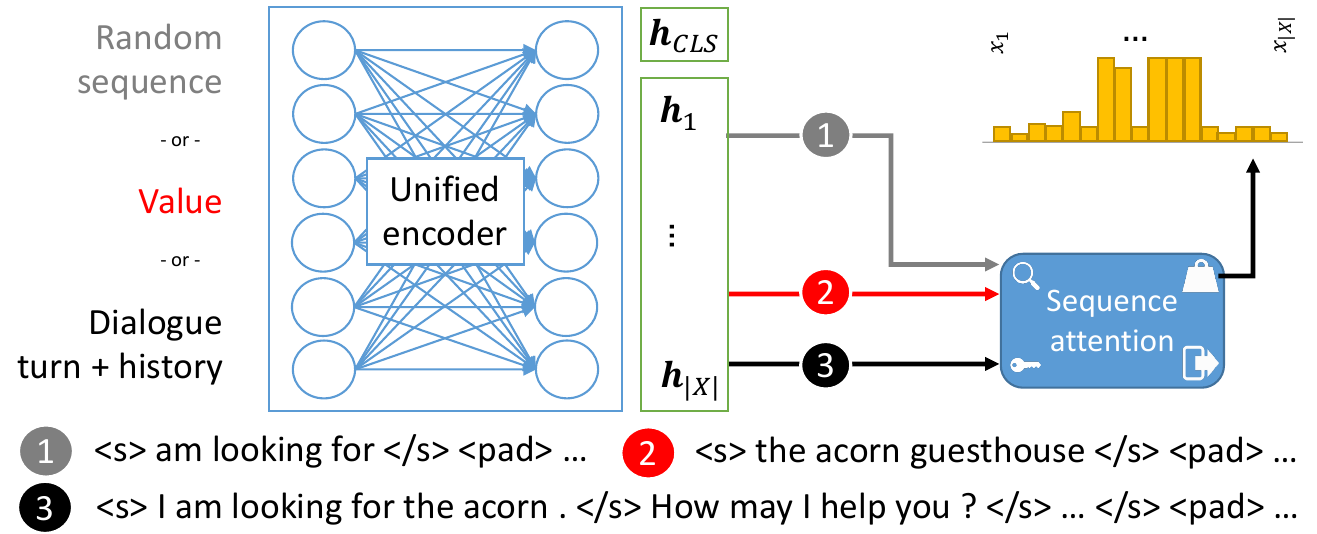}
	\caption{The proto-DST model for value tagging.}
	\label{fig:spanless}
\end{figure}

The proto-DST model consists of the unified encoder and the sequence attention of TripPy-R, as depicted in Figure~\ref{fig:spanless}. Let $D_t = (U_t, M_t)$ be the input to the model, which is encoded as $\mtrx{R}'_t = \mathrm{Enc_S}(x_\mathrm{CLS} \oplus x_\mathrm{NONE} \oplus x_\mathrm{SEP} \oplus U_t \oplus x_\mathrm{SEP} \oplus M_t \oplus x_\mathrm{SEP})$. Let $Y \in D_t$ be a sub-sequence of tokens that was randomly picked from the input, encoded as $\vctr{r}_Y = \mathrm{Enc_P}(x_\mathrm{CLS} \oplus Y \oplus x_\mathrm{SEP})$. In Figure~\ref{fig:spanless}, this corresponds to input types \bettertextcircled{1} and \bettertextcircled{3}. The sequence tagger is then described as
\begin{equation*}
    [\vctr{q}'_{\mathrm{o}}, \vctr{q}'_{\mathrm{w}}] = \mathrm{MHA}_\mathrm{q}(\vctr{r}_{Y}, \mtrx{R}'_{t}, \mtrx{R}'_{t}),
\end{equation*}
analogous to Eq.~(\ref{eq:seq_att}).
For training, we minimise
\begin{equation*}
    \mathcal{L}_{\mathrm{q}} = \ell(\vctr{q}'_{\mathrm{w}}, \vctr{l}^\mathrm{q}_{Y}/\|\vctr{l}^\mathrm{q}_{Y}\|_1), \quad \vctr{l}^\mathrm{q}_{Y} \in \{0, 1\}^{|X|},
\end{equation*}
analogous to Eq.~(\ref{eq:loss}). At each training step, a random negative sample $\bar{Y} \not\in D_t$ rather than a positive sample is picked for training with probability $p_\mathrm{neg}$. For the $Y \in D_t$, the label $\vctr{l}^\mathrm{q}_{Y}$ marks the positions of all tokens of $Y$ in $D_t$. For the $\bar{Y} \not\in D_t$, the label $\vctr{l}^\mathrm{q}_{\bar{Y}}$ puts a weight of 1 onto special token $x_\mathrm{NONE}$ and 0 everywhere else. The desired behaviour of this model is therefore to distribute the maximum amount of the probability mass uniformly among all tokens that belong to the randomly picked sequence.
In case a queried sequence is absent from the input, all probability mass should be assigned to $x_\mathrm{NONE}$. Table~\ref{tab:tagging_example} lists positive and negative training examples.

In order to tag value occurrences in dialogue turns for training with spanless labels, we predict for each value in $\mathit{DS}_t$ its position in $D_t$, given the proto-DST.
Let $s_t^i$ be the value label for slot $S_i$ in turn $t$, which is encoded as $\vctr{r}_{s_t^i} = \mathrm{Enc_P}(x_\mathrm{CLS} \oplus s_t^i \oplus x_\mathrm{SEP})$. Value tagging is performed as
\begin{equation*}
    [\vctr{q}'_{\mathrm{o}}, \vctr{q}'_{\mathrm{w}}] = \mathrm{MHA}_\mathrm{q}(\vctr{r}_{s_t^i}, \mtrx{R}'_{t}, \mtrx{R}'_{t}, \hat{\vctr{r}}_t), \forall s_t^i \in \mathit{DS}_t,
\end{equation*}
which corresponds to input types \bettertextcircled{2} and \bettertextcircled{3} in Figure~\ref{fig:spanless}. $\vctr{q}'_{\mathrm{w}}$ is normalised according to Eq.~(\ref{eq:seq_att_quant}). Table~\ref{tab:tagging_example} shows examples of value tagging with the proto-DST. A set of tag weights $\hat{\vctr{q}}'_{\mathrm{w}}$ is accepted if more than half the probability mass is assigned to word tokens rather than $x_\mathrm{NONE}$. We use a morphological closing operation~\cite{serra1982image} to smooth the tags, i.e.,
\begin{equation}
    \hat{\vctr{q}}'_{\mathrm{w}} \bullet \omega = \delta_{\text{>}\nu}(\hat{\vctr{q}}'_{\mathrm{w}} \smalloplus \omega) \smallominus \omega,
    \label{eq:dae}
\end{equation}
where $\smalloplus$ and $\smallominus$ are the dilation and erosion operators,
$\delta$ is an indicator function, $\hat{\vctr{q}}'_{\mathrm{w}}$ is interpreted as an array, $\omega = [1, 1, 1]$ is a kernel, and $\nu$ is a threshold parameter that allows filtering of tags based on their predicted weights.

\begin{table*}[t]
    \setlength{\tabcolsep}{1.5pt}
    \centering
    \footnotesize
    \begin{tabular}{@{} p{95pt} p{66pt} cccccccccccccc@{}}
    \toprule
    \textbf{Training}        & PMUL1188 & $x_{\mathrm{CLS}}$ & $x_{\mathrm{NONE}}$ & $x_{\mathrm{SEP}}$ & I & \textcolor{blue}{need} & \textcolor{blue}{a} & \textcolor{blue}{train} & to & leave & from & Cambridge & after & 15:30 &  $x_{\mathrm{SEP}}$ \\
    $\textcolor{blue}{Y} \in D_t$              & labels & 0 & 0 & 0 & 0 & 1 & 1 & 1 & 0 & 0 & 0 & 0 & 0 & 0 & 0 \\
    $\bar{Y} \not\in D_t$    & labels & 0 & 1 & 0 & 0 & 0 & 0 & 0 & 0 & 0 & 0 & 0 & 0 & 0 & 0 \\
    \midrule
    \end{tabular}
    \begin{tabular}{@{} p{95pt} p{50pt} ccccccccccccccc@{}}
    \textbf{Tagging}         & PMUL2340 & $x_{\mathrm{CLS}}$ & $x_{\mathrm{NONE}}$ & $x_{\mathrm{SEP}}$ & Hi, & I & am & looking & for & an & \textcolor{blue}{upscale} & restaurant & in & the & \textcolor{blue}{centre} & $x_{\mathrm{SEP}}$ \\
    \textit{rest.-price=expensive}  & prediction & 0 & .25 & 0 & 0 & 0 & 0 & 0 & 0 & 0 & .87 & 0 & 0 & 0 & 0 & 0 \\
    \textit{rest.-area=centre}      & prediction & 0 & 0 & 0 & 0 & 0 & 0 & 0 & 0 & 0 & 0 & 0 & 0 & 0 & .99 & 0 \\
    \bottomrule
    \end{tabular}
    \caption{Top: Training samples for the proto-DST. $Y=\{\text{``need''}, \text{``a''}, \text{``train''}\}$ is a randomly picked sub-sequence in $D_t$. The model needs to tag all tokens belonging to $Y$. For any random sequence $\bar{Y} \not\in D_t$, all probability mass should be assigned to $x_\mathrm{NONE}$. Bottom: Example of tagging the training data with a proto-DST given only spanless labels. The model needs to tag all tokens belonging to the respective values. Note how the proto-DST successfully tagged the word `upscale' as an occurrence of the canonical value \textit{restaurant-price=expensive}.}
    \label{tab:tagging_example}
\end{table*}

Contextual representations enable our value tagger to also identify positions of value variants, i.e., different expressions of the same value (see Table~\ref{tab:tagging_example} for an example).
We tag turns without their history. To generate labels for the history portion, we simply concatenate the tags of the preceding turns with the tags of the current turn.

\subsection{Robustness to Sample Sparsity}

We propose new forms of input-level dropout to increase variance in training samples while preventing an increase in data and training time.

\paragraph{Token Noising}

Targeted feature dropout~\cite{xu2014targeted} has already been used successfully in the form of slot value dropout (SVD) to stabilise DST model training~\cite{chao2019bert, heck-etal-2020-trippy}. During training, SVD replaces tokens of extractable values in their context by a special token $x_{\mathrm{UNK}}$ with a certain probability. The representation of $x_{\mathrm{UNK}}$ amalgamates the contextual representations of all tokens that are not in the encoder's vocabulary $V_{\mathrm{enc}}$ and therefore carries little semantic meaning.

Instead of randomly replacing target tokens with $x_{\mathrm{UNK}}$, we use random tokens from a frequency-sorted $V_{\mathrm{enc}}$. Specifically, a target token is replaced with probability $p_{\mathrm{tn}}$ by a token $x_k \in V_{\mathrm{enc}}$, where $k$ is drawn from a uniform distribution $\mathcal{U}(1, K)$. Since the least frequent tokens in $V_{\mathrm{enc}}$ tend to be nonsensical, we use a cut-off $K \ll |V_{\mathrm{enc}}|$ for $k$. The idea behind this \emph{token noising} is to avoid a train-test discrepancy. With SVD, $x_{\mathrm{UNK}}$ is occasionally presented as target during training, but the model will always encounter valid tokens during inference. With token noising, this mismatch does not occur. Further, token noising increases the variety of observed training samples, while SVD potentially produces duplicate inputs by masking with a placeholder.

\paragraph{History Dropout}

We propose history dropout as another measure to prevent over-fitting due to sample sparsity. With probability $p_{\mathrm{hd}}$, we discard parts of the turn history $H_t$ during training. The cut-off is sampled from $\mathcal{U}(1, t-1)$.
Utilising dialogue history is essential for competitive DST~\cite{heck-etal-2020-trippy}. However, models might learn correlations from sparse samples that do not hold true on new data. The idea of history dropout is to prevent the model from over-relying on the history so as to not be thrown off by previously unencountered conversational styles or contents.

\subsection{Robustness to Unseen Values}
\label{sec:robustness:ssec:values}

Robustness to unseen values is the result of multiple design choices. The applied triple copy strategy as proposed by~\citet{heck-etal-2020-trippy} facilitates value independence. Our proposed token noising and history dropout prevent memorisation of reoccurring patterns.
TripPy-R's value matching provides an alternative prediction for the DS update, in case candidate values are available during inference.
Our model is equipped with the partial masking functionality~\cite{heck-etal-2020-trippy}. Masking may be applied to informed values in the system utterances $M_t,\dots,M_1$ using $x_{\mathrm{UNK}}$, which forces the model to focus on the system utterances' context information rather than specific mentions of values.

\subsection{Robustness to Unseen Slots and Domains}
\label{sec:robustness:ssec:domains}

Domain transfer has the highest demand for generalisability and robustness.
A transfer of the strong triple copy strategy DST baseline to new topics post facto is not possible due to ontology dependence of slot gates, span prediction heads, inform memory and classification heads for coreference resolution. The latter two mechanisms in particular contribute to robustness of DST towards unseen values within known domains~\cite{heck-etal-2020-trippy}. However, the proposed TripPy-R architecture is absolutely vital to establish robustness of triple copy strategy DST to unseen slots across new domains.
TripPy-R is designed to be entirely domain-agnostic by using a model architecture whose parts can be conditioned on natural language descriptions of concepts.

\begin{table*}[t]
  \centering
  \footnotesize
  \begin{tabular}{@{}lcccccccc@{}}
    \toprule
    \multirow{2}{*}{Models} & \multicolumn{2}{c}{sim-M} & \multicolumn{2}{c}{sim-R}  & \multicolumn{2}{c}{WOZ 2.0} & \multicolumn{2}{c}{MultiWOZ 2.1} \\
     & average & best & average & best & average & best & average & best \\
    \midrule
    TripPy (baseline)    & 88.7$\pm$2.7 & 94.0 & 90.4$\pm$1.0 & 91.5 & \textbf{92.3}$\pm$0.6 & \textbf{93.1} & 55.3$\pm$0.9 & 56.3 \\
    \midrule
    TripPy-R w/o value matching & 95.1$\pm$0.9 & 96.1 & 92.0$\pm$0.9 & 93.8 & 91.3$\pm$1.2 & 92.2 & 54.2$\pm$0.2 & 54.3 \\
    \midrule
    TripPy-R & \textbf{95.6}$\pm$1.0 & \textbf{96.8} & 92.3$\pm$2.7 & \textbf{96.2} & 91.5$\pm$0.6 & 92.6 & \textbf{56.0}$\pm$0.3 & \textbf{56.4} \\
    \quad w/o History dropout & 95.4$\pm$0.5 & 96.1 & \textbf{93.2}$\pm$0.9 & 94.7 & 91.6$\pm$1.0 & 93.0 & 55.5$\pm$0.6 & 56.2 \\
    \quad\quad w/o Token noising & 88.6$\pm$3.6 & 94.4 & 92.7$\pm$1.2 & 94.9 & 91.3$\pm$0.7 & 92.5 & 54.8$\pm$0.4 & 55.3 \\
    \quad\quad\quad w/o Joint components & 87.2$\pm$3.9 & 92.6 & 90.8$\pm$0.9 & 91.9 & 91.7$\pm$0.6 & 92.8 & 54.9$\pm$0.3 & 55.3 \\
    \midrule
    TripPy-R w/ spanless training & 95.2$\pm$0.8 & 96.0 & 92.0$\pm$1.5 & 94.0 & 91.1$\pm$0.8 & 92.4 & 55.1$\pm$0.5 & 55.7 \\
    \quad w/o value matching & 92.0$\pm$1.4 & 93.6 & 91.6$\pm$1.3 & 94.5 & 89.0$\pm$0.7 & 90.0 & 51.4$\pm$0.4 & 51.9 \\
    \quad w/ variants    & / & / & / & / & / & / & 55.2$\pm$0.1 & 55.3 \\
    \bottomrule
  \end{tabular}
  \caption{DST results in JGA ($\pm$ denotes standard deviation). \emph{w/o value matching} refers to training and inference.}
  \label{tab:results1}
\end{table*}

\section{Experimental Setup}
\label{sec:experiments}

\subsection{Datasets}

We use MultiWOZ 2.1~\cite{eric2019multiwoz}, WOZ 2.0~\cite{wen2016network}, sim-M and sim-R~\cite{shah2018building} for robustness tests.
MultiWOZ 2.1 is a standard benchmark for multi-domain dialogue modeling that contains 10000+ dialogues covering 5 domains (train, restaurant, hotel, taxi, attraction) and 30 unique domain-slot pairs.
The other datasets are significantly smaller, making sample sparsity an issue.
We test TripPy-R's value independence on two specialised MultiWOZ test sets, $\mathrm{OOO}_{\mathrm{Heck}}$~\cite{heck-etal-2020-trippy} and $\mathrm{OOO}_{\mathrm{Qian}}$~\cite{qian2021annotation}, which replace many values with out-of-ontology (OOO) values.
In addition to MultiWOZ version 2.1, we test TripPy-R on 2.0, 2.2, 2.3 and 2.4~\cite{budzianowski2018multiwoz,zang2020multiwoz,han2020multiwoz,ye2021multiwoz}.

\subsection{Evaluation}

We use joint goal accuracy (JGA) as primary metric to compare between models. The JGA given a test set is the ratio of dialogue turns for which all slots were filled with the correct value (including $\mathit{none}$).
For domain-transfer tests, we report per-domain JGA, and for out-of-ontology prediction experiments, we also report per-slot accuracy.
We repeat each experiment 10 times for small datasets, and three times for MultiWOZ
and report averaged numbers and maximum performance.
For evaluation, we follow~\citet{heck-etal-2020-trippy}.

\subsection{Training}

We initialise our unified encoder with RoBERTa-base~\cite{liu2019roberta}.
The input sequence length is 180 after WordPiece tokenization~\cite{wu2016google}. The loss weights are $(\lambda_{\mathrm{g}}, \lambda_{\mathrm{q}}, \lambda_{\mathrm{f}}, \lambda_{\mathrm{m}}) = (0.8, \frac{1-\lambda_{\mathrm{g}}}{2}, \frac{1-\lambda_{\mathrm{g}}}{2}, 0.1)$.
$\ell_{\mathrm{g}}, \ell_{\mathrm{f}}$ are cross entropy loss, and $\ell_{\mathrm{q}}, \ell_{\mathrm{m}}$ are mean squared error loss.
We use Adam optimiser~\cite{kingma2014adam} and back-propagate through the entire network including the encoder.
We also back-propagate the error for slot encodings, since we re-encode them at every step. The learning rate is 5e-5 after a warmup portion of 10\% (5\% for MultiWOZ), then decays linearly. The maximum number of epochs is 20 for MultiWOZ, 50 for WOZ 2.0 and 100 for sim-M/R. We use early stopping with patience (20\% of max. epoch), based on the development set JGA. The batch size is 16 (32 for MultiWOZ).
During training, the encoder output dropout rate is 30\%, and $p_\mathrm{tn}$ $=$ $p_\mathrm{hd}$ $=$ $30\%$
(10\% for MultiWOZ). The weight decay rate is 0.01. For token noising, we set $K = 0.2\cdot|V_{\mathrm{enc}}|$. We weight $\ell_{\mathrm{g}}$ for $\mathit{none}$ cases with 0.1. For value matching we tune $\tau$ in decrements of 0.1 on the development sets.

For spanless training, the maximum length of random token sequences for the proto-DST model training is 4.
The maximum number of epochs is 50 for the WOZ datasets and 100 for sim-M/R. The negative sampling probability is $p_\mathrm{neg} = 10\%$.

\section{Experimental Results}

\subsection{Learning from Spanless Labels}

\begin{figure}[t]
	\centering
	\includegraphics[page=2, trim=6.9cm 6.5cm 9.5cm 5.55cm, clip=true, width=1.00\linewidth,]{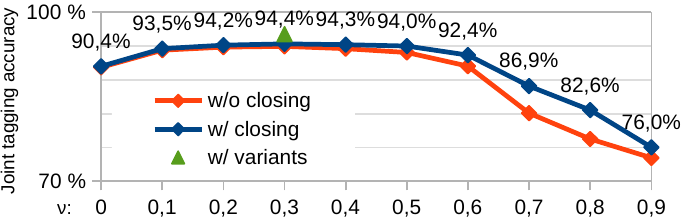}
	\caption{Tagging performance of the proto-DST model depending on the weight threshold $\nu$.}
	\label{fig:dae}
	\vspace{-10pt}
\end{figure}

The quality of the proto-DST for value tagging determines whether or not training without explicit span labels leads to useful DST models. We evaluate the tagging performance on the example of MultiWOZ 2.1 by calculating the ratio of turns for which all tokens are assigned the correct ``\emph{IO}'' tag.
Figure~\ref{fig:dae} plots the joint tagging accuracy across slots, dependent on the weight threshold in Eq.~(\ref{eq:dae}). It can be seen that an optimal threshold is $\nu = 0.3$. We found this to be true across all datasets. We also found that the morphological closing operation generally improves tagging accuracy.
Typical errors that are corrected by this post-processing are gaps caused by occasionally failing to tag special characters within values, e.g. ``\texttt{:}'' in times with \texttt{hh:mm} format, and imprecisions caused by insecurities of the model when tagging long and complex values such as movie names.
Average tagging accuracy across slots is 99.8\%. This is particularly noteworthy since values in MultiWOZ can be expressed with a wide variety (e.g., "expensive" might be expressed as "upscale", "fancy", and so on). We attribute the high tagging accuracy to the expressiveness of the encoder-generated semantic contextual representations.

Table~\ref{tab:results1} lists the JGA of TripPy-R when trained without manual span labels. For the small datasets we did not use $x_\mathrm{NONE}$ and negative sampling, as it did not make a significant difference.
We see that performance is on par with models that were trained with full supervision. If value matching on top of sequence tagging is not used, performance is slightly below its supervised counterparts. We observed that value matching compensates for minor errors caused by the sequence tagger that was trained on automatic labels.\footnote{Note that training with value matching also affects the training of the sequence tagger, be it with or without using span labels.}

\paragraph{Impact of Tagging Variants}

While our proto-DST model already achieves very high accuracy on all slots including the ones that expect values with many variants, we tested whether explicit tagging of variants may further improve performance. For instance, if a turn contains the (canonical) value ``expensive'' for slot \emph{hotel-pricerange}, but expressed as ``upscale'',
we would explicitly tag such variants.
While this strategy further improved the joint tagging accuracy from 94.4\% to 96.1\% (Figure~\ref{fig:dae}), we did not see a rise in DST performance (Table~\ref{tab:results1}). In other words, the contextual encoder is powerful enough to endow the proto-DST model with the ability to tag variants of values, based on semantic similarity, which renders any extra supervision for this task unnecessary.

\subsection{Handling Sample Sparsity}

\paragraph{Impact of Token Noising}

We experienced that traditional SVD leads to performance gains on sim-M, but not on any of the other tested datasets, confirming~\citet{heck-etal-2020-trippy}.
In contrast, token noising improved the JGA for sim-M/R considerably. Note that in Table~\ref{tab:results1}, the TripPy baseline for sim-M already uses SVD.
On MultiWOZ 2.1, we observed minor improvements. As with SVD, WOZ 2.0 remained unaffected. The ontology for WOZ 2.0 is rather limited and remains the same for training and testing. This is not the case for the other datasets, where values occur during testing that were never seen during training.
By all appearances, presenting the model with a more diverse set of dropped-out training examples helps generalisation more than using a single placeholder token. This seems especially true when there are only few value candidates per slot, and few training samples to learn from. A particularly exemplaric case is found in the sim-M dataset. Without token noising, trained models regularly end up interpreting the value ``last christmas'' as \textit{movie-date} rather than \textit{movie-name}, based on its semantic similarity to dates. Token noising on the other hand forces the model to put more emphasis on context rather than token identities, which effectively removes the occurrence of this error.

\paragraph{Impact of History Dropout}

\begin{figure}[t]
	\centering
	\includegraphics[page=4, trim=8.8cm 6.2cm 8.7cm 6cm, clip=true, width=1.00\linewidth,]{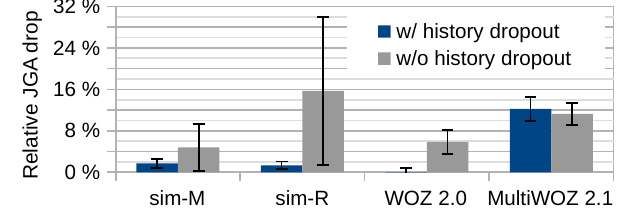}
	\caption{Performance loss due to mismatched training and testing conditions. Here, history is provided during training, but not during testing. sim-M/R and WOZ 2.0 show clear signs of over-fitting without history dropout.}
	\label{fig:hd}
	\vspace{-10pt}
\end{figure}

Table~\ref{tab:results1} shows that history dropout does not adversely affect DST performance.
This is noteworthy since utilising the full dialogue history is the standard in contemporary works due to its importance for adequate tracking.
History dropout effectively reduces the amount of training data by omitting parts of the model input. At the same time training samples are diversified,
preventing the model from memorising patterns in the dialogue history and promoting generalisation.
Figure~\ref{fig:hd} shows the severe effects of over-fitting to the dialogue history on small datasets, when not using history dropout. Here, models were only provided the current turn as input, without historical context. Models with history dropout fare considerably better, showing that they do not over-rely on the historical information. Models without history dropout do not only perform much worse, their performance is also extremely unstable. On sim-R, the span from least to highest relative performance drop is 0\% to 39.4\%. The results on MultiWOZ point to the importance of the historical information for proper tracking on more challenging scenarios. Here, performance drops equally in the absence of dialogue history, whether or not history dropout was used.

\subsection{Handling Unseen Values}
\label{sec:experiments:ssec:oov}

We probed value independence on two out-of-ontology test sets for MultiWOZ. $\mathrm{OOO}_\mathrm{Heck}$ replaces most values by fictional but still meaningful values that are not in the original ontology. Replacements are consistent, i.e., the same value is always replaced by the same fictional stand-in. The overall OOO rate is 84\%. $\mathrm{OOO}_\mathrm{Qian}$ replaces only values of slots that expect names (i.e., \emph{name}, \emph{departure} and \emph{destination}) with values from a different ontology. Replacements are not consistent across dialogues, and such that names are shared across all slots, e.g., street names may become hotel names, restaurants may become train stops and so on, i.e., the distinction between concepts is lost.

\begin{table}[t]
  \setlength{\tabcolsep}{9pt}
  \centering
  \footnotesize
  \begin{tabular}{@{}lcc@{}}
    \toprule
    Models & $\mathrm{OOO}_{\mathrm{Heck}}$ & $\mathrm{OOO}_{\mathrm{Qian}}$ \\
    \midrule
    TripPy & 40.1$\pm$1.9 & 29.2$\pm$1.9 \\
    \citet{qian2021annotation} & / & 27.0$\pm$2.0 \\
    \midrule
    TripPy-R & 42.2$\pm$0.8 & 29.7$\pm$0.7 \\
    TripPy-R + masking & \textbf{43.0}$\pm$1.5 & \textbf{36.0}$\pm$1.6 \\
    \bottomrule
  \end{tabular}
  \caption{Performance in JGA on artificial out-of-ontology test sets ($\pm$ denotes standard deviation).}
  \label{tab:ooo}
  \vspace{-5pt}
\end{table}

\begin{table}[t]
  \setlength{\tabcolsep}{3pt}
  \centering
  \footnotesize
  \begin{tabular}{@{}lcccccc@{}}
    \toprule
    \multirow{2}{*}{Models} & \multicolumn{5}{c}{Domains} \\
    & hotel & rest. & attr. & train & taxi & avg. \\
    \midrule
    TRADE~\citeyearpar{wu2019transferable,campagna2020zero} & 19.5 & 16.4 & 22.8 & 22.9 & 59.2 & 28.2 \\
    MA-DST~\citeyearpar{kumar2020ma} & 16.3 & 13.6 & 22.5 & 22.8 & 59.3 & 26.9 \\
    SUMBT~\citeyearpar{lee2019sumbt,campagna2020zero} & \textbf{19.8} & 16.5 & 22.6 & 22.5 & 59.5 & 28.2 \\
    \citet{li2021zero}$^*$   & 18.5 & \textbf{21.1} & 23.7 & \textbf{24.3} & 59.1 & \textbf{29.3} \\
    \midrule
    TripPy-R                & 18.3 & 15.3 & \textbf{27.1} & 23.7 & \textbf{61.5} & 29.2 \\
    \midrule\midrule
    \citet{li2021zero}$^{**}$& 24.4 & 26.2 & 31.3 & 29.1 & 59.6 & 34.1 \\
    \bottomrule
  \end{tabular}
  \caption{Best zero-shot DST results for various models on MultiWOZ 2.1 in JGA. $^*$ \citet{li2021zero} presents considerably higher numbers for models with data augmentation. We compare against a model without data augmentation. $^{**}$ is a model with three times as many parameters as ours.}
  \label{tab:zero}
  \vspace{-10pt}
\end{table}

Table~\ref{tab:ooo} lists the results. The performance loss is more graceful on $\mathrm{OOO}_\mathrm{Heck}$, and we see that TripPy-R has an advantage over TripPy.
The performance drop is more severe on $\mathrm{OOO}_\mathrm{Qian}$, with comparable JGA to the baseline of~\citet{qian2021annotation}, which is a generative model. The authors of that work attribute the performance degradation to hallucinations caused by memorisation effects.
For our extractive model the main reason is found in the slot gate. The relative slot gate performance drop for the \emph{train} domain-slots is 23.3\%, while for other domain-slots it is 6.4\%. We believe the reason is that most of the arbitrary substitutes carry no characteristics of train stops, but of other domains instead. This is less of a problem for the \emph{taxi} domain for instance, since taxi stops are of a variety of location types.
The issue of value-to-domain mismatch can be mitigated somewhat with informed value masking in system utterances (Section~\ref{sec:robustness:ssec:values}).
While this does not particularly affect our model on the regular test set or on the more domain-consistent $\mathrm{OOO}_\mathrm{Heck}$, we can see much better generalisation on $\mathrm{OOO}_\mathrm{Qian}$.

\subsection{Handling Unseen Slots and Domains}
\label{sec:experiments:ssec:zeroshot}

Table~\ref{tab:results1} shows that moving from slot specific to slot independent components only marginally affects DST performance, while enabling tracking of dialogues with unseen domains and slots.

\paragraph{Zero-shot Performance}

We conducted zero-shot experiments on MultiWOZ 2.1 by excluding all dialogues of a domain $d$ from training and then evaluating the model on dialogues of $d$. In Table~\ref{tab:zero}, we compare TripPy-R to recent models that support slot independence. Even though we did not specifically optimise TripPy-R for zero-shot abilities, our model shows a level of robustness that is competitive with other contemporary methods.

\begin{figure}[t]
	\centering
	\includegraphics[page=3, trim=8cm 5.1cm 7.8cm 5.2cm, clip=true, width=1.00\linewidth,]{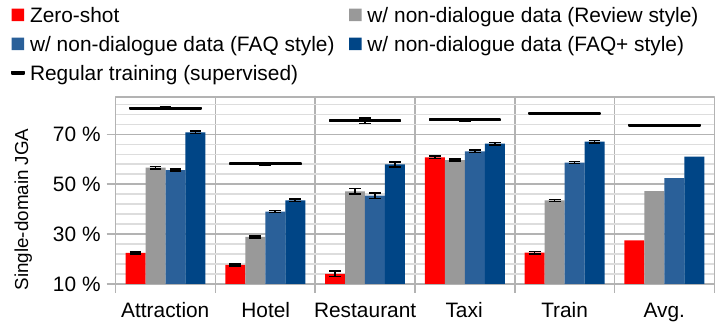}
	\caption{Performance of TripPy-R after training with non-dialogue style data from a held-out domain.}
	\label{fig:nondiag}
	\vspace{-5pt}
\end{figure}

\paragraph{Impact of Non-dialogue Data}

Besides zero-shot abilities, we were curious, is it feasible to improve dialogue state tracking by learning the required mechanics purely from non-dialogue data? This is a non-trivial task, as the model needs to generalise knowledge learned from unstructured data to dialogue, i.e., sequences of alternating system and user utterances.
We conducted this experiment by converting MultiWOZ dialogues of a held-out domain $d$ into non-dialogue format for training. For $d$, the model only sees isolated sentences or sentence pairs, i.e., without any structure of a dialogue. Consequently, there is no ``turn'' history from which the model could learn. The assumption is that one would have some way to label sequences of interest in non-dialogue sentences, for instance with a semantic parser. As this is a feasibility study, we resort to the slot labels in $DS_t$, which simulates having labels of very high accuracy.
We tested three different data formats, 
\begin{enumerate*}[(1)]
    \item \emph{Review style}: Only system utterances with statements are used to learn from;
    \item \emph{FAQ style}: A training example is formed by a user question and the following system answer. Note that this is contrary to what TripPy-R naturally expects, which is a querying system and a responding user; and
    \item \emph{FAQ+ style}: Combines review and FAQ style examples and adds user questions again as separate examples.
\end{enumerate*}

Figure~\ref{fig:nondiag} shows that we observed considerable improvements across all held-out domains when using non-dialogue data to learn from. Learning from additional data, even if unstructured, is particularly beneficial for unique slots, such as the \textit{restaurant-food} slot which the model can not learn about from any other domain in MultiWOZ (as is reflected in a poor zero-shot performance as well).
We also found that learning benefits from the combination of different formats. 
The heightened performance given the FAQ+ style data is not an effect of more data, but of its presentation, since we mainly \emph{re-use} inputs with different formats.
This observation is reminiscent of findings in psychology. \citet{horst2011get} showed that children benefited from being read the same story repeatedly. Furthermore, \citet{johns2016influence} showed that contextual diversity positively affects word learning in adults.
Note that this kind of learning is in contrast to few-shot learning and leveraging artificial dialogue data, which either require fine-grained manual labels or high-level knowledge of how dialogues are structured. Even though the data we used
is far-removed from what a dialogue state tracker expects, TripPy-R still manages to learn how to appropriately track these new domains.

\begin{figure}[t]
	\centering
	\includegraphics[page=1, trim=7.55cm 5.5cm 6.1cm 4.9cm, clip=true, width=1.00\linewidth,]{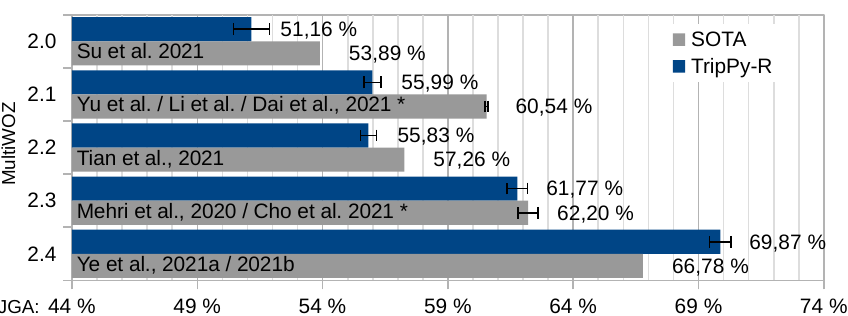}
	\caption{Comparison of TripPy-R and SOTA open vocabulary DST models. $^*$ denotes TripPy-style models.}
	\label{fig:mwoz}
	\vspace{-5pt}
\end{figure}

\subsection{Performance Comparison}

We evaluated on five versions of MultiWOZ to place TripPy-R among contemporary work. Versions 2.1 and 2.2 mainly propose general corrections to the labels of MultiWOZ 2.0. Version 2.3 unifies annotations between dialogue acts and dialogue states. In contrast, version 2.4 removes all values that were mentioned by the system from the dialogue state, unless they are proper names. Figure~\ref{fig:mwoz} plots the results. The performance of TripPy-R is considerably better on 2.3 and 2.4. This can be attributed to a more accurate prediction of the \emph{inform} cases due to better test ground truths.

For fairness, we restricted our comparison to models that have the same general abilities, i.e., they ought to be open-vocabulary and without data-specific architectures. The SOTA on 2.0~\cite{su2021multi} proposes a unified generative dialogue model to solve multiple tasks including DST and benefits from pre-training on various dialogue corpora. While profiting from more data in general, its heterogeneity in particular did not affect DST performance. \citet{yu2020score}, \citet{li2020coco} and \citet{dai2021preview} currently share the top of the leaderboard for 2.1, all of which propose TripPy-style models that leverage data augmentation. The main reason for performance improvements lies in the larger amount of data and in diversifying samples. TripPy-R does not rely on more data, but diversifies training samples with token noising and history dropout. On 2.2, the method of \citet{tian-etal-2021-amendable} performs best with a two-pass generative approach that utilises an error recovery mechanism. This mechanism can correct generation errors such as caused by hallucination, which is a phenomenon that does not occur with TripPy-R. However, their error recovery also has the potential to avoid propagation of errors made early in the dialogue, which is demonstrated by a heightened performance. \citet{cho2021checkdst} report numbers for the method of \citet{mehri2020dialoglue} on 2.3, which is another TripPy-style model using an encoder that was pre-trained on millions of conversations, thus greatly benefiting from specialised knowledge. For 2.4, the current SOTA with the properties as stated above is presented by~\citet{ye2021slot} and reported in~\citet{ye2021multiwoz}, which is now surpassed by TripPy-R. The major difference to our model is the use of slot self-attention, which allows their model to learn correlations between slot occurrences. While TripPy-R does not model slot correlations directly, it does however explicitly learn to resolve coreferences.

\subsection{Implications of the Results}

The zero-shot capabilities of our proposed TripPy-R model open the door to many new applications. However, it should be noted that its performance on an unseen arbitrary domain and on unseen arbitrary slots will likely degrade. In such cases it would be more appropriate to perform adaptation, which the TripPy-R framework facilitates. This means that one would transfer the model as presented in Sections~\ref{sec:robustness:ssec:values} and~\ref{sec:robustness:ssec:domains} and continue fine-tuning with limited---and potentially unstructured (see Section~\ref{sec:experiments:ssec:zeroshot})---data from the new domain. Nonetheless, in applications such as e-commerce~\cite{zhang-etal-2018-modeling} or customer support~\cite{garcia-sardina-etal-2018-es}, whenever new slots or even domains are introduced, they are to a great extent related to ones that a deployed system is familiar with. We believe that the zero-shot performance presented in Table~\ref{tab:zero} is highly indicative of this set-up, as MultiWOZ domains are different, yet to some extent related.

Further, the TripPy-R model facilitates future applications in complex domains such as healthcare. One of the biggest obstacles to harnessing large amounts of natural language data in healthcare is the required labelling effort. This is particularly the case for applications in psychology, as can be seen from the recent work of~\citet{rojas-barahona-etal-2018-deep} where only 5K out of 1M interactions where labelled with spans for so called \emph{thinking-errors} by physiologists. A framework like TripPy-R can completely bypass this step by utilising its proto-DST, as presented in Section~\ref{sec:robustness:ssec:spanless}, eliminating the overbearing labelling effort.

\section{Conclusion}

In this work we present methods to facilitate robust extractive dialogue state tracking with weak supervision and sparse data.
Our proposed architecture---TripPy-R---utilises a unified encoder, the attention mechanism and conditioning on natural language descriptions of concepts to facilitate parameter sharing and zero-shot transfer. We leverage similarity based value matching as an optional step after value extraction, without violating the principle of ontology independence.

We demonstrated the feasibility of training without manual span labels using a self-trained proto-DST model. Learning from spanless labels enables us to leverage data with weaker supervision. We showed that token noising and history dropout mitigate issues of pattern memorisation and train-test discrepancies. We achieved competitive zero-shot performance and demonstrated in a feasibility study that TripPy-R can learn to track new domains by learning from non-dialogue data. We achieve either competitive or state-of-the-art performance on all tested benchmarks.
For future work we continue to investigate learning from non-dialogue data, potentially in a continuous fashion over the lifetime of a dialogue system.

\section*{Acknowledgements}

We thank the anonymous reviewers and the action editors for their valuable feedback. M. Heck, N. Lubis, C. van Niekerk and S. Feng are supported by funding provided by the Alexander von Humboldt Foundation in the framework of the Sofja Kovalevskaja Award endowed by the Federal Ministry of Education and Research, while C. Geishauser and H-C. Lin are supported by funds from the European Research Council (ERC) provided under the Horizon 2020 research and innovation programme (Grant agreement No. STG2018804636). Computing resources were provided by Google Cloud and HHU ZIM.

\bibliography{tacl2018}

\begin{thebibliography}{64}
\expandafter\ifx\csname natexlab\endcsname\relax\def\natexlab#1{#1}\fi

\bibitem[{Ba et~al.(2016)Ba, Kiros, and Hinton}]{ba2016layer}
Jimmy~Lei Ba, Jamie~Ryan Kiros, and Geoffrey~E. Hinton. 2016.
\newblock \href {http://arxiv.org/abs/1607.06450} {Layer normalization}.
\newblock \emph{CoRR}, abs/1607.06450v1.

\bibitem[{Bahdanau et~al.(2015)Bahdanau, Cho, and
  Bengio}]{DBLP:journals/corr/BahdanauCB14}
Dzmitry Bahdanau, Kyunghyun Cho, and Yoshua Bengio. 2015.
\newblock \href {http://arxiv.org/abs/1409.0473v7} {Neural machine translation
  by jointly learning to align and translate}.
\newblock In \emph{Proceedings of the 3rd International Conference on Learning
  Representations}, San Diego, CA, USA.

\bibitem[{Bapna et~al.(2017)Bapna, T{\"u}r, Hakkani-T{\"u}r, and
  Heck}]{bapna2017towards}
Ankur Bapna, Gokhan T{\"u}r, Dilek Hakkani-T{\"u}r, and Larry Heck. 2017.
\newblock \href {https://doi.org/10.21437/Interspeech.2017-518} {Towards
  zero-shot frame semantic parsing for domain scaling}.
\newblock In \emph{Proceedings of Interspeech 2017}, pages 2476--2480.

\bibitem[{Budzianowski et~al.(2018)Budzianowski, Wen, Tseng, Casanueva, Ultes,
  Ramadan, and Ga{\v{s}}i{\'c}}]{budzianowski2018multiwoz}
Pawe{\l} Budzianowski, Tsung-Hsien Wen, Bo-Hsiang Tseng, I{\~n}igo Casanueva,
  Stefan Ultes, Osman Ramadan, and Milica Ga{\v{s}}i{\'c}. 2018.
\newblock \href {https://doi.org/10.18653/v1/D18-1547} {{M}ulti{WOZ} - {A}
  large-scale multi-domain {W}izard-of-{O}z dataset for task-oriented dialogue
  modelling}.
\newblock In \emph{Proceedings of the 2018 Conference on Empirical Methods in
  Natural Language Processing}, pages 5016--5026, Brussels, Belgium.
  Association for Computational Linguistics.

\bibitem[{Campagna et~al.(2020)Campagna, Foryciarz, Moradshahi, and
  Lam}]{campagna2020zero}
Giovanni Campagna, Agata Foryciarz, Mehrad Moradshahi, and Monica Lam. 2020.
\newblock \href {https://doi.org/10.18653/v1/2020.acl-main.12} {Zero-shot
  transfer learning with synthesized data for multi-domain dialogue state
  tracking}.
\newblock In \emph{Proceedings of the 58th Annual Meeting of the Association
  for Computational Linguistics}, pages 122--132, Online. Association for
  Computational Linguistics.

\bibitem[{Chao and Lane(2019)}]{chao2019bert}
Guan-Lin Chao and Ian Lane. 2019.
\newblock \href {https://doi.org/10.21437/Interspeech.2019-1355} {{BERT-DST}:
  {S}calable end-to-end dialogue state tracking with bidirectional encoder
  representations from transformer}.
\newblock In \emph{Proceedings of Interspeech 2019}, pages 1468--1472.

\bibitem[{Cho et~al.(2021)Cho, Sankar, Lin, Sadagopan, Shayandeh, Celikyilmaz,
  May, and Beirami}]{cho2021checkdst}
Hyundong Cho, Chinnadhurai Sankar, Christopher Lin, Kaushik~Ram Sadagopan,
  Shahin Shayandeh, Asli Celikyilmaz, Jonathan May, and Ahmad Beirami. 2021.
\newblock \href {http://arxiv.org/abs/2112.08321} {{CheckDST}: {M}easuring
  real-world generalization of dialogue state tracking performance}.
\newblock \emph{CoRR}, abs/2112.08321v1.

\bibitem[{Clark and Brennan(1991)}]{clark1991grounding}
Herbert~H. Clark and Susan~E. Brennan. 1991.
\newblock \href {https://doi.org/10.1037/10096-006} {Grounding in
  communication}.
\newblock In Lauren~B. Resnick, John~M. Levine, and Stephanie~D. Teasley,
  editors, \emph{Perspectives on Socially Shared Cognition}, pages 127--149.
  American Psychological Association, Washington, USA.

\bibitem[{Dai et~al.(2021)Dai, Li, Li, Sun, Huang, Si, and
  Zhu}]{dai2021preview}
Yinpei Dai, Hangyu Li, Yongbin Li, Jian Sun, Fei Huang, Luo Si, and Xiaodan
  Zhu. 2021.
\newblock \href {https://doi.org/10.18653/v1/2021.acl-short.111} {Preview,
  attend and review: {S}chema-aware curriculum learning for multi-domain
  dialogue state tracking}.
\newblock In \emph{Proceedings of the 59th Annual Meeting of the Association
  for Computational Linguistics and the 11th International Joint Conference on
  Natural Language Processing (Volume 2: Short Papers)}, pages 879--885,
  Online. Association for Computational Linguistics.

\bibitem[{Deriu et~al.(2021)Deriu, Rodrigo, Otegi, Echegoyen, Rosset, Agirre,
  and Cieliebak}]{deriu2021survey}
Jan Deriu, Alvaro Rodrigo, Arantxa Otegi, Guillermo Echegoyen, Sophie Rosset,
  Eneko Agirre, and Mark Cieliebak. 2021.
\newblock \href {https://doi.org/10.1007/s10462-020-09866-x} {Survey on
  evaluation methods for dialogue systems}.
\newblock \emph{Artificial Intelligence Review}, 54(1):755--810.

\bibitem[{Devlin et~al.(2019)Devlin, Chang, Lee, and
  Toutanova}]{devlin2018bert}
Jacob Devlin, Ming-Wei Chang, Kenton Lee, and Kristina Toutanova. 2019.
\newblock \href {https://doi.org/10.18653/v1/N19-1423} {{BERT}: {P}re-training
  of deep bidirectional transformers for language understanding}.
\newblock In \emph{Proceedings of the 2019 Conference of the North {A}merican
  Chapter of the Association for Computational Linguistics: Human Language
  Technologies, Volume 1 (Long and Short Papers)}, pages 4171--4186,
  Minneapolis, Minnesota. Association for Computational Linguistics.

\bibitem[{Edlund et~al.(2008)Edlund, Gustafson, Heldner, and
  Hjalmarsson}]{edlund2008towards}
Jens Edlund, Joakim Gustafson, Mattias Heldner, and Anna Hjalmarsson. 2008.
\newblock \href {https://doi.org/10.1016/j.specom.2008.04.002} {Towards
  human-like spoken dialogue systems}.
\newblock \emph{Speech Communication}, 50(8-9):630--645.

\bibitem[{Eric et~al.(2020)Eric, Goel, Paul, Sethi, Agarwal, Gao, Kumar, Goyal,
  Ku, and Hakkani-T{\"u}r}]{eric2019multiwoz}
Mihail Eric, Rahul Goel, Shachi Paul, Abhishek Sethi, Sanchit Agarwal, Shuyang
  Gao, Adarsh Kumar, Anuj Goyal, Peter Ku, and Dilek Hakkani-T{\"u}r. 2020.
\newblock \href {https://aclanthology.org/2020.lrec-1.53} {{M}ulti{WOZ} 2.1:
  {A} consolidated multi-domain dialogue dataset with state corrections and
  state tracking baselines}.
\newblock In \emph{Proceedings of the 12th Language Resources and Evaluation
  Conference}, pages 422--428, Marseille, France. European Language Resources
  Association.

\bibitem[{Gao et~al.(2019)Gao, Sethi, Agarwal, Chung, and
  Hakkani-T{\"u}r}]{gao2019dialog}
Shuyang Gao, Abhishek Sethi, Sanchit Agarwal, Tagyoung Chung, and Dilek
  Hakkani-T{\"u}r. 2019.
\newblock \href {https://doi.org/10.18653/v1/W19-5932} {Dialog state tracking:
  {A} neural reading comprehension approach}.
\newblock In \emph{Proceedings of the 20th Annual Meeting of the Special
  Interest Group on Discourse and Dialogue}, pages 264--273, Stockholm, Sweden.
  Association for Computational Linguistics.

\bibitem[{Garc{\'\i}a-Sardi{\~n}a et~al.(2018)Garc{\'\i}a-Sardi{\~n}a, Serras,
  and del Pozo}]{garcia-sardina-etal-2018-es}
Laura Garc{\'\i}a-Sardi{\~n}a, Manex Serras, and Arantza del Pozo. 2018.
\newblock \href {https://aclanthology.org/L18-1125} {{ES}-{P}ort: {A}
  spontaneous spoken human-human technical support corpus for dialogue research
  in {S}panish}.
\newblock In \emph{Proceedings of the 11th International Conference on Language
  Resources and Evaluation}, pages 781--785, Miyazaki, Japan. European Language
  Resources Association.

\bibitem[{Han et~al.(2021)Han, Liu, Takanabu, Lian, Huang, Wan, Peng, and
  Huang}]{han2020multiwoz}
Ting Han, Ximing Liu, Ryuichi Takanabu, Yixin Lian, Chongxuan Huang, Dazhen
  Wan, Wei Peng, and Minlie Huang. 2021.
\newblock \href {https://doi.org/10.1007/978-3-030-88483-3_16} {Multi{WOZ} 2.3:
  {A} multi-domain task-oriented dialogue dataset enhanced with annotation
  corrections and co-reference annotation}.
\newblock In \emph{CCF International Conference on Natural Language Processing
  and Chinese Computing}, pages 206--218. Springer.

\bibitem[{Heck et~al.(2020{\natexlab{a}})Heck, Geishauser, Lin, Lubis, Moresi,
  van Niekerk, and Ga{\v{s}}i{\'c}}]{heck2020out}
Michael Heck, Christian Geishauser, Hsien-chin Lin, Nurul Lubis, Marco Moresi,
  Carel van Niekerk, and Milica Ga{\v{s}}i{\'c}. 2020{\natexlab{a}}.
\newblock \href {https://doi.org/10.18653/v1/2020.coling-main.596} {Out-of-task
  training for dialog state tracking models}.
\newblock In \emph{Proceedings of the 28th International Conference on
  Computational Linguistics}, pages 6767--6774, Barcelona, Spain (Online).
  International Committee on Computational Linguistics.

\bibitem[{Heck et~al.(2020{\natexlab{b}})Heck, van Niekerk, Lubis, Geishauser,
  Lin, Moresi, and Ga{\v{s}}i{\'c}}]{heck-etal-2020-trippy}
Michael Heck, Carel van Niekerk, Nurul Lubis, Christian Geishauser, Hsien-Chin
  Lin, Marco Moresi, and Milica Ga{\v{s}}i{\'c}. 2020{\natexlab{b}}.
\newblock \href {https://aclanthology.org/2020.sigdial-1.4} {{T}rip{P}y: {A}
  triple copy strategy for value independent neural dialog state tracking}.
\newblock In \emph{Proceedings of the 21st Annual Meeting of the Special
  Interest Group on Discourse and Dialogue}, pages 35--44, 1st virtual meeting.
  Association for Computational Linguistics.

\bibitem[{Hendrycks and Gimpel(2016)}]{hendrycks2016gaussian}
Dan Hendrycks and Kevin Gimpel. 2016.
\newblock \href {http://arxiv.org/abs/1606.08415} {Gaussian error linear units
  ({GELU}s)}.
\newblock \emph{CoRR}, abs/1606.08415v4.

\bibitem[{Horst et~al.(2011)Horst, Parsons, and Bryan}]{horst2011get}
Jessica~S. Horst, Kelly~L. Parsons, and Natasha~M. Bryan. 2011.
\newblock \href {https://doi.org/10.3389/fpsyg.2011.00017} {Get the story
  straight: {C}ontextual repetition promotes word learning from storybooks}.
\newblock \emph{Frontiers in Psychology}, 2:17.

\bibitem[{Johns et~al.(2016)Johns, Dye, and Jones}]{johns2016influence}
Brendan~T. Johns, Melody Dye, and Michael~N. Jones. 2016.
\newblock \href {https://doi.org/10.3758/s13423-015-0980-7} {The influence of
  contextual diversity on word learning}.
\newblock \emph{Psychonomic Bulletin \& Review}, 23(4):1214--1220.

\bibitem[{Kim et~al.(2020)Kim, Yang, Kim, and Lee}]{kim2019efficient}
Sungdong Kim, Sohee Yang, Gyuwan Kim, and Sang-Woo Lee. 2020.
\newblock \href {https://doi.org/10.18653/v1/2020.acl-main.53} {Efficient
  dialogue state tracking by selectively overwriting memory}.
\newblock In \emph{Proceedings of the 58th Annual Meeting of the Association
  for Computational Linguistics}, pages 567--582, Online. Association for
  Computational Linguistics.

\bibitem[{Kingma and Ba(2015)}]{kingma2014adam}
Diederik~P. Kingma and Jimmy Ba. 2015.
\newblock \href {http://arxiv.org/abs/1412.6980v9} {Adam: {A} method for
  stochastic optimization}.
\newblock In \emph{Proceedings of the 3rd International Conference on Learning
  Representations}, San Diego, CA, USA.

\bibitem[{Kumar et~al.(2020)Kumar, Ku, Goyal, Metallinou, and
  Hakkani-T{\"u}r}]{kumar2020ma}
Adarsh Kumar, Peter Ku, Anuj~Kumar Goyal, Angeliki Metallinou, and Dilek
  Hakkani-T{\"u}r. 2020.
\newblock \href {https://doi.org/10.1609/aaai.v34i05.6322} {{MA-DST}:
  {M}ulti-attention based scalable dialog state tracking}.
\newblock In \emph{Proceedings of the 34th AAAI Conference on Artificial
  Intelligence}, volume~34, pages 8107--8114.

\bibitem[{Larsson and Traum(2000)}]{larsson2000information}
Staffan Larsson and David~R. Traum. 2000.
\newblock \href {https://doi.org/10.1017/S1351324900002539} {Information state
  and dialogue management in the {TRINDI} dialogue move engine toolkit}.
\newblock \emph{Natural Language Engineering}, 6(3-4):323--340.

\bibitem[{Lee et~al.(2019)Lee, Lee, and Kim}]{lee2019sumbt}
Hwaran Lee, Jinsik Lee, and Tae-Yoon Kim. 2019.
\newblock \href {https://doi.org/10.18653/v1/P19-1546} {{SUMBT}:
  {S}lot-utterance matching for universal and scalable belief tracking}.
\newblock In \emph{Proceedings of the 57th Annual Meeting of the Association
  for Computational Linguistics}, pages 5478--5483, Florence, Italy.
  Association for Computational Linguistics.

\bibitem[{Li et~al.(2020)Li, Yavuz, Hashimoto, Li, Niu, Rajani, Yan, Zhou, and
  Xiong}]{li2020coco}
Shiyang Li, Semih Yavuz, Kazuma Hashimoto, Jia Li, Tong Niu, Nazneen Rajani,
  Xifeng Yan, Yingbo Zhou, and Caiming Xiong. 2020.
\newblock \href {https://openreview.net/forum?id=eom0IUrF__F} {{CoCo}:
  {C}ontrollable counterfactuals for evaluating dialogue state trackers}.
\newblock In \emph{International Conference on Learning Representations}.

\bibitem[{Li et~al.(2021)Li, Cao, Sridhar, Zhu, Li, Hamza, and
  McAuley}]{li2021zero}
Shuyang Li, Jin Cao, Mukund Sridhar, Henghui Zhu, Shang-Wen Li, Wael Hamza, and
  Julian McAuley. 2021.
\newblock \href {https://doi.org/10.18653/v1/2021.eacl-main.91} {Zero-shot
  generalization in dialog state tracking through generative question
  answering}.
\newblock In \emph{Proceedings of the 16th Conference of the European Chapter
  of the Association for Computational Linguistics: Main Volume}, pages
  1063--1074, Online. Association for Computational Linguistics.

\bibitem[{Liang et~al.(2021)Liang, Poddar, and Szarvas}]{liang2021attention}
Shuailong Liang, Lahari Poddar, and Gyuri Szarvas. 2021.
\newblock \href {http://arxiv.org/abs/2101.11958} {Attention guided dialogue
  state tracking with sparse supervision}.
\newblock \emph{CoRR}, abs/2101.11958v1.

\bibitem[{Lin et~al.(2021)Lin, Liu, Madotto, Moon, Zhou, Crook, Wang, Yu, Cho,
  Subba, and Fung}]{lin2021zero}
Zhaojiang Lin, Bing Liu, Andrea Madotto, Seungwhan Moon, Zhenpeng Zhou, Paul
  Crook, Zhiguang Wang, Zhou Yu, Eunjoon Cho, Rajen Subba, and Pascale Fung.
  2021.
\newblock \href {https://doi.org/10.18653/v1/2021.emnlp-main.622} {Zero-shot
  dialogue state tracking via cross-task transfer}.
\newblock In \emph{Proceedings of the 2021 Conference on Empirical Methods in
  Natural Language Processing}, pages 7890--7900, Online and Punta Cana,
  Dominican Republic. Association for Computational Linguistics.

\bibitem[{Liu and Lane(2017)}]{liu2017end}
Bing Liu and Ian Lane. 2017.
\newblock \href {https://doi.org/10.21437/Interspeech.2017-1326} {An end-to-end
  trainable neural network model with belief tracking for task-oriented
  dialog}.
\newblock In \emph{Proceedings of Interspeech 2017}, pages 2506--2510.

\bibitem[{Liu et~al.(2019)Liu, Ott, Goyal, Du, Joshi, Chen, Levy, Lewis,
  Zettlemoyer, and Stoyanov}]{liu2019roberta}
Yinhan Liu, Myle Ott, Naman Goyal, Jingfei Du, Mandar Joshi, Danqi Chen, Omer
  Levy, Mike Lewis, Luke Zettlemoyer, and Veselin Stoyanov. 2019.
\newblock \href {http://arxiv.org/abs/1907.11692} {Ro{BERT}a: {A} robustly
  optimized {BERT} pretraining approach}.
\newblock \emph{CoRR}, abs/1907.11692v1.

\bibitem[{Ma et~al.(2019)Ma, Zeng, Zhu, Li, Yang, Yao, Zhou, and
  Shen}]{ma2019end}
Yue Ma, Zengfeng Zeng, Dawei Zhu, Xuan Li, Yiying Yang, Xiaoyuan Yao, Kaijie
  Zhou, and Jianping Shen. 2019.
\newblock \href {http://arxiv.org/abs/1912.09297} {An end-to-end dialogue state
  tracking system with machine reading comprehension and wide {\&} deep
  classification}.
\newblock \emph{CoRR}, abs/1912.09297v2.

\bibitem[{Mehri et~al.(2020)Mehri, Eric, and
  Hakkani{-}T{\"{u}}r}]{mehri2020dialoglue}
Shikib Mehri, Mihail Eric, and Dilek Hakkani{-}T{\"{u}}r. 2020.
\newblock \href {http://arxiv.org/abs/2009.13570} {Dialo{GLUE}: {A} natural
  language understanding benchmark for task-oriented dialogue}.
\newblock \emph{CoRR}, abs/2009.13570v1.

\bibitem[{Mrk{\v{s}}i{\'c} et~al.(2017)Mrk{\v{s}}i{\'c}, {\'O}~S{\'e}aghdha,
  Wen, Thomson, and Young}]{mrkvsic2016neural}
Nikola Mrk{\v{s}}i{\'c}, Diarmuid {\'O}~S{\'e}aghdha, Tsung-Hsien Wen, Blaise
  Thomson, and Steve Young. 2017.
\newblock \href {https://doi.org/10.18653/v1/P17-1163} {Neural belief tracker:
  {D}ata-driven dialogue state tracking}.
\newblock In \emph{Proceedings of the 55th Annual Meeting of the Association
  for Computational Linguistics (Volume 1: Long Papers)}, pages 1777--1788,
  Vancouver, Canada. Association for Computational Linguistics.

\bibitem[{Namazifar et~al.(2021)Namazifar, Papangelis, T{\"u}r, and
  Hakkani-T{\"u}r}]{namazifar2021language}
Mahdi Namazifar, Alexandros Papangelis, Gokhan T{\"u}r, and Dilek
  Hakkani-T{\"u}r. 2021.
\newblock \href {https://doi.org/10.1109/ICASSP39728.2021.9413810} {Language
  model is all you need: {N}atural language understanding as question
  answering}.
\newblock In \emph{Proceedings of the 2021 IEEE International Conference on
  Acoustics, Speech and Signal Processing}, pages 7803--7807. IEEE.

\bibitem[{Ni et~al.(2021)Ni, Young, Pandelea, Xue, Adiga, and
  Cambria}]{ni2021recent}
Jinjie Ni, Tom Young, Vlad Pandelea, Fuzhao Xue, Vinay Adiga, and Erik Cambria.
  2021.
\newblock \href {http://arxiv.org/abs/2105.04387} {Recent advances in deep
  learning based dialogue systems: {A} systematic survey}.
\newblock \emph{CoRR}, abs/2105.04387v4.

\bibitem[{Nouri and Hosseini-Asl(2018)}]{nouri2018toward}
Elnaz Nouri and Ehsan Hosseini-Asl. 2018.
\newblock \href {https://doi.org/10.48550/arXiv.1812.00899v1} {Toward scalable
  neural dialogue state tracking model}.
\newblock In \emph{Proceedings of the 2nd Conversational AI workshop at the
  32nd Conference on Neural Information Processing Systems}, Montréal, Canada.

\bibitem[{Qian et~al.(2021)Qian, Beirami, Lin, De, Geramifard, Yu, and
  Sankar}]{qian2021annotation}
Kun Qian, Ahmad Beirami, Zhouhan Lin, Ankita De, Alborz Geramifard, Zhou Yu,
  and Chinnadhurai Sankar. 2021.
\newblock \href {https://aclanthology.org/2021.sigdial-1.35} {Annotation
  inconsistency and entity bias in {M}ulti{WOZ}}.
\newblock In \emph{Proceedings of the 22nd Annual Meeting of the Special
  Interest Group on Discourse and Dialogue}, pages 326--337, Singapore and
  Online. Association for Computational Linguistics.

\bibitem[{Ramshaw and Marcus(1995)}]{ramshaw-marcus-1995-text}
Lance Ramshaw and Mitch Marcus. 1995.
\newblock \href {https://aclanthology.org/W95-0107} {Text chunking using
  transformation-based learning}.
\newblock In \emph{Proceedings of the 3rd Workshop on Very Large Corpora},
  pages 82--94.

\bibitem[{Rastogi et~al.(2020{\natexlab{a}})Rastogi, Zang, Sunkara, Gupta, and
  Khaitan}]{rastogi2020schema}
Abhinav Rastogi, Xiaoxue Zang, Srinivas Sunkara, Raghav Gupta, and Pranav
  Khaitan. 2020{\natexlab{a}}.
\newblock \href {http://arxiv.org/abs/2002.01359} {Schema-guided dialogue state
  tracking task at {DSTC8}}.
\newblock \emph{CoRR}, abs/2002.01359v1.

\bibitem[{Rastogi et~al.(2020{\natexlab{b}})Rastogi, Zang, Sunkara, Gupta, and
  Khaitan}]{rastogi2020towards}
Abhinav Rastogi, Xiaoxue Zang, Srinivas Sunkara, Raghav Gupta, and Pranav
  Khaitan. 2020{\natexlab{b}}.
\newblock \href {https://doi.org/10.1609/aaai.v34i05.6394} {Towards scalable
  multi-domain conversational agents: {T}he schema-guided dialogue dataset}.
\newblock In \emph{Proceedings of the 34th AAAI Conference on Artificial
  Intelligence}, volume~34, pages 8689--8696.

\bibitem[{Ren et~al.(2018)Ren, Xie, Chen, and Yu}]{ren2018towards}
Liliang Ren, Kaige Xie, Lu~Chen, and Kai Yu. 2018.
\newblock \href {https://doi.org/10.18653/v1/D18-1299} {Towards universal
  dialogue state tracking}.
\newblock In \emph{Proceedings of the 2018 Conference on Empirical Methods in
  Natural Language Processing}, pages 2780--2786, Brussels, Belgium.
  Association for Computational Linguistics.

\bibitem[{Rojas-Barahona et~al.(2018)Rojas-Barahona, Tseng, Dai, Mansfield,
  Ramadan, Ultes, Crawford, and
  Ga{\v{s}}i{\'c}}]{rojas-barahona-etal-2018-deep}
Lina~M. Rojas-Barahona, Bo-Hsiang Tseng, Yinpei Dai, Clare Mansfield, Osman
  Ramadan, Stefan Ultes, Michael Crawford, and Milica Ga{\v{s}}i{\'c}. 2018.
\newblock \href {https://doi.org/10.18653/v1/W18-5606} {Deep learning for
  language understanding of mental health concepts derived from cognitive
  behavioural therapy}.
\newblock In \emph{Proceedings of the 9th International Workshop on Health Text
  Mining and Information Analysis}, pages 44--54, Brussels, Belgium.
  Association for Computational Linguistics.

\bibitem[{Serra(1982)}]{serra1982image}
Jean Serra. 1982.
\newblock \href {https://dl.acm.org/doi/10.5555/1098652} {\emph{Image Analysis
  and Mathematical Morphology}}.
\newblock Academic Press, Inc.

\bibitem[{Shah et~al.(2018)Shah, Hakkani{-}T{\"{u}}r, T{\"{u}}r, Rastogi,
  Bapna, Nayak, and Heck}]{shah2018building}
Pararth Shah, Dilek Hakkani{-}T{\"{u}}r, G{\"{o}}khan T{\"{u}}r, Abhinav
  Rastogi, Ankur Bapna, Neha Nayak, and Larry~P. Heck. 2018.
\newblock \href {http://arxiv.org/abs/1801.04871} {Building a conversational
  agent overnight with dialogue self-play}.
\newblock \emph{CoRR}, abs/1801.04871v1.

\bibitem[{Su et~al.(2022)Su, Shu, Mansimov, Gupta, Cai, Lai, and
  Zhang}]{su2021multi}
Yixuan Su, Lei Shu, Elman Mansimov, Arshit Gupta, Deng Cai, Yi-An Lai, and
  Yi~Zhang. 2022.
\newblock \href {https://doi.org/10.18653/v1/2022.acl-long.319} {Multi-task
  pre-training for plug-and-play task-oriented dialogue system}.
\newblock In \emph{Proceedings of the 60th Annual Meeting of the Association
  for Computational Linguistics (Volume 1: Long Papers)}, pages 4661--4676,
  Dublin, Ireland. Association for Computational Linguistics.

\bibitem[{Tian et~al.(2021)Tian, Huang, Lin, Bao, He, Yang, Wu, Wang, and
  Sun}]{tian-etal-2021-amendable}
Xin Tian, Liankai Huang, Yingzhan Lin, Siqi Bao, Huang He, Yunyi Yang, Hua Wu,
  Fan Wang, and Shuqi Sun. 2021.
\newblock \href {https://doi.org/10.18653/v1/2021.nlp4convai-1.8} {Amendable
  generation for dialogue state tracking}.
\newblock In \emph{Proceedings of the 3rd Workshop on Natural Language
  Processing for Conversational AI}, pages 80--92, Online. Association for
  Computational Linguistics.

\bibitem[{Vaswani et~al.(2017)Vaswani, Shazeer, Parmar, Uszkoreit, Jones,
  Gomez, Kaiser, and Polosukhin}]{vaswani2017attention}
Ashish Vaswani, Noam Shazeer, Niki Parmar, Jakob Uszkoreit, Llion Jones,
  Aidan~N. Gomez, {\L}ukasz Kaiser, and Illia Polosukhin. 2017.
\newblock \href
  {https://proceedings.neurips.cc/paper/2017/file/3f5ee243547dee91fbd053c1c4a845aa-Paper.pdf}
  {Attention is all you need}.
\newblock In \emph{Advances in Neural Information Processing Systems},
  volume~30, pages 5998--6008. Curran Associates, Inc.

\bibitem[{Wen et~al.(2017)Wen, Vandyke, Mrk{\v{s}}i{\'c}, Ga{\v{s}}i{\'c},
  Rojas-Barahona, Su, Ultes, and Young}]{wen2016network}
Tsung-Hsien Wen, David Vandyke, Nikola Mrk{\v{s}}i{\'c}, Milica
  Ga{\v{s}}i{\'c}, Lina~M. Rojas-Barahona, Pei-Hao Su, Stefan Ultes, and Steve
  Young. 2017.
\newblock \href {https://aclanthology.org/E17-1042} {A network-based end-to-end
  trainable task-oriented dialogue system}.
\newblock In \emph{Proceedings of the 15th Conference of the {E}uropean Chapter
  of the Association for Computational Linguistics: Volume 1, Long Papers},
  pages 438--449, Valencia, Spain. Association for Computational Linguistics.

\bibitem[{Wu et~al.(2020)Wu, Hoi, Socher, and Xiong}]{wu2020tod}
Chien-Sheng Wu, Steven~C.H. Hoi, Richard Socher, and Caiming Xiong. 2020.
\newblock \href {https://doi.org/10.18653/v1/2020.emnlp-main.66} {{TOD}-{BERT}:
  {P}re-trained natural language understanding for task-oriented dialogue}.
\newblock In \emph{Proceedings of the 2020 Conference on Empirical Methods in
  Natural Language Processing}, pages 917--929, Online. Association for
  Computational Linguistics.

\bibitem[{Wu et~al.(2019)Wu, Madotto, Hosseini-Asl, Xiong, Socher, and
  Fung}]{wu2019transferable}
Chien-Sheng Wu, Andrea Madotto, Ehsan Hosseini-Asl, Caiming Xiong, Richard
  Socher, and Pascale Fung. 2019.
\newblock \href {https://doi.org/10.18653/v1/P19-1078} {Transferable
  multi-domain state generator for task-oriented dialogue systems}.
\newblock In \emph{Proceedings of the 57th Annual Meeting of the Association
  for Computational Linguistics}, pages 808--819, Florence, Italy. Association
  for Computational Linguistics.

\bibitem[{Wu et~al.(2016)Wu, Schuster, Chen, Le, Norouzi, Macherey, Krikun,
  Cao, Gao, Macherey, Klingner, Shah, Johnson, Liu, Kaiser, Gouws, Kato, Kudo,
  Kazawa, Stevens, Kurian, Patil, Wang, Young, Smith, Riesa, Rudnick, Vinyals,
  Corrado, Hughes, and Dean}]{wu2016google}
Yonghui Wu, Mike Schuster, Zhifeng Chen, Quoc~V. Le, Mohammad Norouzi, Wolfgang
  Macherey, Maxim Krikun, Yuan Cao, Qin Gao, Klaus Macherey, Jeff Klingner,
  Apurva Shah, Melvin Johnson, Xiaobing Liu, {\L}ukasz Kaiser, Stephan Gouws,
  Yoshikiyo Kato, Taku Kudo, Hideto Kazawa, Keith Stevens, George Kurian,
  Nishant Patil, Wei Wang, Cliff Young, Jason Smith, Jason Riesa, Alex Rudnick,
  Oriol Vinyals, Greg Corrado, Macduff Hughes, and Jeffrey Dean. 2016.
\newblock \href {http://arxiv.org/abs/1609.08144} {Google's neural machine
  translation system: {B}ridging the gap between human and machine
  translation}.
\newblock \emph{CoRR}, abs/1609.08144v2.

\bibitem[{Xu and Hu(2018)}]{xu2018end}
Puyang Xu and Qi~Hu. 2018.
\newblock \href {https://doi.org/10.18653/v1/P18-1134} {An end-to-end approach
  for handling unknown slot values in dialogue state tracking}.
\newblock In \emph{Proceedings of the 56th Annual Meeting of the Association
  for Computational Linguistics (Volume 1: Long Papers)}, pages 1448--1457,
  Melbourne, Australia. Association for Computational Linguistics.

\bibitem[{Xu and Sarikaya(2014)}]{xu2014targeted}
Puyang Xu and Ruhi Sarikaya. 2014.
\newblock \href
  {https://www.isca-speech.org/archive_v0/archive_papers/interspeech_2014/i14_0258.pdf}
  {Targeted feature dropout for robust slot filling in natural language
  understanding}.
\newblock In \emph{Fifteenth Annual Conference of the International Speech
  Communication Association}, pages 258--262.

\bibitem[{Ye et~al.(2021{\natexlab{a}})Ye, Manotumruksa, and
  Yilmaz}]{ye2021multiwoz}
Fanghua Ye, Jarana Manotumruksa, and Emine Yilmaz. 2021{\natexlab{a}}.
\newblock \href {http://arxiv.org/abs/2104.00773} {Multi{WOZ} 2.4: {A}
  multi-domain task-oriented dialogue dataset with essential annotation
  corrections to improve state tracking evaluation}.
\newblock \emph{CoRR}, abs/2104.00773v1.

\bibitem[{Ye et~al.(2021{\natexlab{b}})Ye, Manotumruksa, Zhang, Li, and
  Yilmaz}]{ye2021slot}
Fanghua Ye, Jarana Manotumruksa, Qiang Zhang, Shenghui Li, and Emine Yilmaz.
  2021{\natexlab{b}}.
\newblock \href {https://doi.org/10.1145/3442381.3449939} {Slot self-attentive
  dialogue state tracking}.
\newblock In \emph{Proceedings of the Web Conference 2021}, pages 1598--1608.

\bibitem[{Young et~al.(2010)Young, Ga{\v{s}}i{\'c}, Keizer, Mairesse,
  Schatzmann, Thomson, and Yu}]{young2010hidden}
Steve Young, Milica Ga{\v{s}}i{\'c}, Simon Keizer, Fran{\c{c}}ois Mairesse,
  Jost Schatzmann, Blaise Thomson, and Kai Yu. 2010.
\newblock \href {https://doi.org/10.1016/j.csl.2009.04.001} {The hidden
  information state model: {A} practical framework for {POMDP}-based spoken
  dialogue management}.
\newblock \emph{Computer Speech \& Language}, 24(2):150--174.

\bibitem[{Yu et~al.(2020)Yu, Zhang, Polozov, Meek, and Awadallah}]{yu2020score}
Tao Yu, Rui Zhang, Alex Polozov, Christopher Meek, and Ahmed~Hassan Awadallah.
  2020.
\newblock \href {https://openreview.net/forum?id=oyZxhRI2RiE} {{SCoRe}:
  {P}re-training for context representation in conversational semantic
  parsing}.
\newblock In \emph{International Conference on Learning Representations}.

\bibitem[{Zang et~al.(2020)Zang, Rastogi, Sunkara, Gupta, Zhang, and
  Chen}]{zang2020multiwoz}
Xiaoxue Zang, Abhinav Rastogi, Srinivas Sunkara, Raghav Gupta, Jianguo Zhang,
  and Jindong Chen. 2020.
\newblock \href {https://doi.org/10.18653/v1/2020.nlp4convai-1.13}
  {{M}ulti{WOZ} 2.2 : {A} dialogue dataset with additional annotation
  corrections and state tracking baselines}.
\newblock In \emph{Proceedings of the 2nd Workshop on Natural Language
  Processing for Conversational AI}, pages 109--117, Online. Association for
  Computational Linguistics.

\bibitem[{Zhang et~al.(2020{\natexlab{a}})Zhang, Hashimoto, Wu, Wang, Yu,
  Socher, and Xiong}]{zhang2019find}
Jianguo Zhang, Kazuma Hashimoto, Chien-Sheng Wu, Yao Wang, Philip Yu, Richard
  Socher, and Caiming Xiong. 2020{\natexlab{a}}.
\newblock \href {https://aclanthology.org/2020.starsem-1.17} {Find or classify?
  {D}ual strategy for slot-value predictions on multi-domain dialog state
  tracking}.
\newblock In \emph{Proceedings of the 9th Joint Conference on Lexical and
  Computational Semantics}, pages 154--167, Barcelona, Spain (Online).
  Association for Computational Linguistics.

\bibitem[{Zhang et~al.(2020{\natexlab{b}})Zhang, Takanobu, Zhu, Huang, and
  Zhu}]{zhang2020recent}
Zheng Zhang, Ryuichi Takanobu, Qi~Zhu, MinLie Huang, and XiaoYan Zhu.
  2020{\natexlab{b}}.
\newblock \href {https://doi.org/10.1007/s11431-020-1692-3} {Recent advances
  and challenges in task-oriented dialog systems}.
\newblock \emph{Science China Technological Sciences}, 63:2011--2027.

\bibitem[{Zhang et~al.(2018)Zhang, Li, Zhu, Zhao, and
  Liu}]{zhang-etal-2018-modeling}
Zhuosheng Zhang, Jiangtong Li, Pengfei Zhu, Hai Zhao, and Gongshen Liu. 2018.
\newblock \href {https://aclanthology.org/C18-1317} {Modeling multi-turn
  conversation with deep utterance aggregation}.
\newblock In \emph{Proceedings of the 27th International Conference on
  Computational Linguistics}, pages 3740--3752, Santa Fe, New Mexico, USA.
  Association for Computational Linguistics.

\bibitem[{Zhong et~al.(2018)Zhong, Xiong, and Socher}]{zhong2018global}
Victor Zhong, Caiming Xiong, and Richard Socher. 2018.
\newblock \href {https://doi.org/10.18653/v1/P18-1135} {Global-locally
  self-attentive encoder for dialogue state tracking}.
\newblock In \emph{Proceedings of the 56th Annual Meeting of the Association
  for Computational Linguistics (Volume 1: Long Papers)}, pages 1458--1467,
  Melbourne, Australia. Association for Computational Linguistics.

\end{thebibliography}
\bibliographystyle{acl_natbib}

\end{document}